\documentclass{article}

\PassOptionsToPackage{numbers, compress}{natbib}

 \usepackage[eandd, final]{neurips_2026}

\usepackage[utf8]{inputenc} 
\usepackage[T1]{fontenc}    
\usepackage{hyperref}       
\usepackage{url}            
\usepackage{booktabs}       
\usepackage{amsfonts}       
\usepackage{nicefrac}       
\usepackage{microtype}      
\usepackage{xcolor}         
\usepackage{amsmath}
\usepackage{graphicx}
\usepackage{multirow}
\usepackage{float}

\newif\ifshowchanges
\showchangestrue

\title{TopoGraphRAG-Bench: Evaluating Multimodal GraphRAG on Layout-Grounded Evidence Reasoning}

\author{%
  \normalfont
  Ruochi Li\textsuperscript{1},
  Jianzhe Lin\textsuperscript{5,\textdagger},
  Haoxuan Zhang\textsuperscript{2},
  Haihua Chen\textsuperscript{3} \\
  Junhua Ding\textsuperscript{4},
  Edward Gehringer\textsuperscript{1,*},
  Yang Zhang\textsuperscript{3,\textdagger,*} \\[0.4em]
  \textsuperscript{1}Department of Computer Science,
  North Carolina State University, USA \\
  \textsuperscript{2}Department of Information Science,
  University of North Texas, USA \\
  \textsuperscript{3}The Anuradha and Vikas Sinha Department of Data Science,
  University of North Texas, USA \\
  \textsuperscript{4}School of Computing,
  University of Wyoming, USA \\
  \textsuperscript{5}Independent Researcher \\[0.3em]
  {\footnotesize
    \textsuperscript{1}\texttt{\{rli14, efg\}@ncsu.edu}
    \quad
    \textsuperscript{2}\texttt{\{haoxuanzhang\}@my.unt.edu}
    \quad
    \textsuperscript{3}\texttt{\{yang.zhang, haihua.chen\}@unt.edu}
  } \\[0.1em]
  {\footnotesize
    \textsuperscript{4}\texttt{\{junhua.ding\}@uwyo.edu}
    \textsuperscript{5}\texttt{\{jianzhelin\}@meta.com}
  }
}

\begin{document}

\maketitle

\begingroup
\renewcommand{\thefootnote}{\fnsymbol{footnote}}
\footnotetext[1]{Corresponding Authors.}
\footnotetext[2]{Project Leader.}
\endgroup

\begin{abstract}
Real-world documents distribute evidence across text, tables, figures, and captions within complex page layouts. Answering complex questions over such documents therefore requires more than retrieving relevant passages: systems must recover the evidence topology that connects heterogeneous evidence units. Existing GraphRAG evaluations remain largely text-centered, while multimodal document RAG benchmarks assess cross-modal retrieval and generation without directly evaluating recovery of the intended evidence topology. We introduce \textsc{TopoGraphRAG-Bench}, a layout-grounded benchmark for multimodal evidence reasoning in GraphRAG, comprising 2,024 questions over 201 long, visually rich documents. Questions are constructed bottom-up from text, figure, and table evidence units under three controlled topologies: single-hop retrieval, bridge-chain reasoning, and multi-source synthesis. To ensure that questions preserve their intended structure, we apply counterfactual validation for shortcut resistance, modality necessity, and evidence necessity. We evaluate text-only GraphRAG, page-level visual retrieval, and multimodal GraphRAG systems using retrieval, generation, and topology-aware reasoning metrics. Multimodal GraphRAG systems achieve the strongest overall performance, but still fail when visual-textual evidence alignment or multi-unit composition is incomplete. Text-only GraphRAG struggles when key dependencies are grounded in figures or tables, while page-level visual retrieval lacks the fine-grained structure needed for topology recovery. These findings motivate GraphRAG systems that move beyond text-derived entity relation graphs to explicitly model document layouts, cross-modal evidence alignment, and the reasoning roles of evidence units. Code and data are available at \url{https://richardlrc.github.io/TopoGraphRAG-Bench/}.

\end{abstract}

\section{Introduction}
Recent advances in multimodal large language models and retrieval-augmented generation have reshaped how intelligent systems interact with complex documents, where evidence is often distributed across prose, tables, figures, charts, captions, and layout structures. Prior work on document visual question answering, multimodal QA, and long-document understanding has shown that answering document questions often requires grounding in textual, tabular, and visual sources across pages and layouts \citep{mathew2021docvqa,talmor2021multimodalqa,ma2024mmlongbench}. However, many information needs go beyond single-fact retrieval from one passage, table cell, or figure caption. Complex questions often require connecting multiple pieces of evidence \citep{trivedi2022musique,tangmultihop}: one unit may reveal an intermediate entity or condition that determines which subsequent unit is relevant, while several units may jointly support a comparison, summary, or higher-level pattern. In these cases, the retrieval objective shifts from finding isolated relevant evidence to recovering the dependency structure among evidence units. We therefore view complex document retrieval as a structure-recovery problem: the system must identify relevant evidence units as well as the entities, relations, and higher-order connections that bind them together.

GraphRAG has emerged as a promising paradigm for this structure-recovery view of retrieval by organizing textual chunks, entities, relations, and summaries into graph structures that support long-range dependency modeling, multi-hop reasoning, and synthesis beyond local similarity search. Existing GraphRAG benchmarks evaluate graph construction, knowledge retrieval, and generation across tasks such as fact retrieval, complex reasoning, contextual summarization, and creative generation \citep{xiang2025use}. However, these evaluations remain largely text-centered: the evidence graph is typically built from textual chunks, entities, or textual summaries. In parallel, document-centric multimodal RAG benchmarks such as MMDocRAG, UniDoc-Bench, and REAL-MM-RAG have advanced multimodal page retrieval, evidence selection, quote selection, and answer generation over text, figures, and tables \citep{dong2025benchmarking,peng2025unidoc,wasserman2025real}. Yet these benchmarks are not designed to diagnose whether systems recover the intended evidence topology across multimodal document layouts. This leaves open the core question for multimodal GraphRAG: whether graph-structured retrieval can recover and support reasoning over layout-grounded multimodal evidence structures, rather than being limited to text-derived entities and chunks.

To study this question, we introduce a layout-grounded benchmark for multimodal evidence reasoning in GraphRAG. Rather than generating complex questions directly from whole documents, we construct them bottom up: atomic questions are generated from layout-grounded evidence units and then composed through explicit evidence links into three reasoning topologies, single-hop retrieval, bridge-chain reasoning, and multi-source synthesis. We further apply counterfactual validation for shortcut resistance, modality necessity, and evidence necessity, so that each instance preserves its intended evidence structure. The resulting benchmark contains 2,024 questions across 201 documents, including 302 single-hop, 1,014 bridge-chain, and 708 synthesis questions, with 773 cross-modal instances. We evaluate text-only GraphRAG, VisRAG-style page retrieval, and multimodal GraphRAG systems using retrieval and generation metrics. Results show that multimodal GraphRAG systems perform best overall, but each paradigm faces distinct limitations: text-only GraphRAG misses visual and tabular dependencies, page-level visual retrieval struggles with fine-grained evidence topology, and multimodal GraphRAG remains limited by modality alignment failures and incomplete synthesis across heterogeneous evidence.

In summary, our contributions are:
\begin{itemize}

    \item \textbf{Topology-aware multimodal GraphRAG benchmark.}
    We introduce \textsc{TopoGraphRAG-Bench}, a layout-grounded benchmark with 2,024 QA pairs over 201 long documents, covering single-hop retrieval, bridge-chain reasoning, and multi-source synthesis across text, figures, and tables.

    \item \textbf{Bottom-up evidence-topology construction.}
    We construct questions from verifiable layout-grounded evidence units rather than whole-document prompts, and use counterfactual validation to enforce shortcut resistance, modality necessity, and evidence necessity.

    \item \textbf{Diagnostic evaluation of GraphRAG paradigms.}
    We evaluate text-only GraphRAG, page-level visual retrieval, and multimodal GraphRAG systems, showing that current systems still struggle with cross-modal alignment and fine-grained evidence-topology recovery.

\end{itemize}

\section{Related Work}
\paragraph{GraphRAG methods and benchmarks.}
GraphRAG extends retrieval augmented generation by organizing information into graph structures over chunks, entities, relations, communities, or summaries, enabling retrieval beyond local similarity search. 
Representative systems include hierarchical or community-based retrieval methods such as RAPTOR and Microsoft GraphRAG, graph-indexed systems such as LightRAG, personalized PageRank-based methods such as HippoRAG, and KG- or GNN-based systems such as G-Retriever, GFM-RAG, and ToG \citep{sarthi2024raptor,edge2024local,guo2024lightrag,gutierrez2024hipporag,he2024g,luogfm,sunthink}. Recent GraphRAG benchmarks evaluate whether graph-based retrieval improves fact retrieval, complex reasoning, contextual summarization, and generation over conventional RAG \citep{xiang2025use,zhou2025depth}. However, these evaluations remain largely text-centred, with graphs constructed from textual chunks, entities, relations, or summaries, leaving multimodal document evidence structures underexplored.

\paragraph{Multimodal document RAG benchmarks.}
Document understanding benchmarks have evolved from single-page DocVQA tasks to long-document and retrieval-augmented settings \citep{mathew2021docvqa,mathew2022infographicvqa,zhu2022towards,tito2023hierarchical,ma2024mmlongbench,zou2025docbench}. Recent multimodal document RAG benchmarks further evaluate retrieval and generation across pages, layouts, figures, tables, and screenshots. MMDocRAG introduces multi-page cross-modal evidence chains, quote selection, and multimodal answer generation; UniDoc Bench compares text-only, image-only, fusion, and joint retrieval paradigms; REAL MM RAG emphasizes realistic multimodal retrieval queries and accurate labeling; and MMDocIR and ViDoRe focus on multimodal document retrieval \citep{dong2025benchmarking,peng2025unidoc,wasserman2025real,dong2025mmdocir,loison2026vidore}. These works provide strong multimodal evaluation settings, but they are not primarily designed to diagnose whether graph structured retrieval can recover evidence topologies across multimodal document layouts.

\paragraph{Multi-evidence reasoning and question construction.}
Multi-hop QA benchmarks such as HotpotQA, 2WikiMultiHopQA, MuSiQue, and MultiHop RAG show that complex questions often require connecting multiple pieces of evidence rather than retrieving a single fact \citep{yang2018hotpotqa,ho2020constructing,trivedi2022musique,tangmultihop}. Multimodal benchmarks extend this idea to text, tables, images, charts, and long documents, including MultiModalQA, CHARGE, BRIDGE, and DocHop QA \citep{talmor2021multimodalqa,yang2025benchmarking,xiang2026bridge,park2025dochop}. Building on these foundations, we construct questions directly from layout-grounded evidence units in complete documents, retaining their page, layout, modality, and reasoning roles. Our benchmark covers both entity-mediated bridge chains and synthesis over 3--6 complementary evidence points, with topology-specific checks of evidence necessity. These annotations link the required reasoning structure to retrievable document evidence, supporting separate evaluation of evidence retrieval and evidence composition in multimodal GraphRAG.

\section{Benchmark}
\label{benchmark_section}
\subsection{Construction}

\begin{figure*}[t]
\centering
\includegraphics[width=\textwidth]{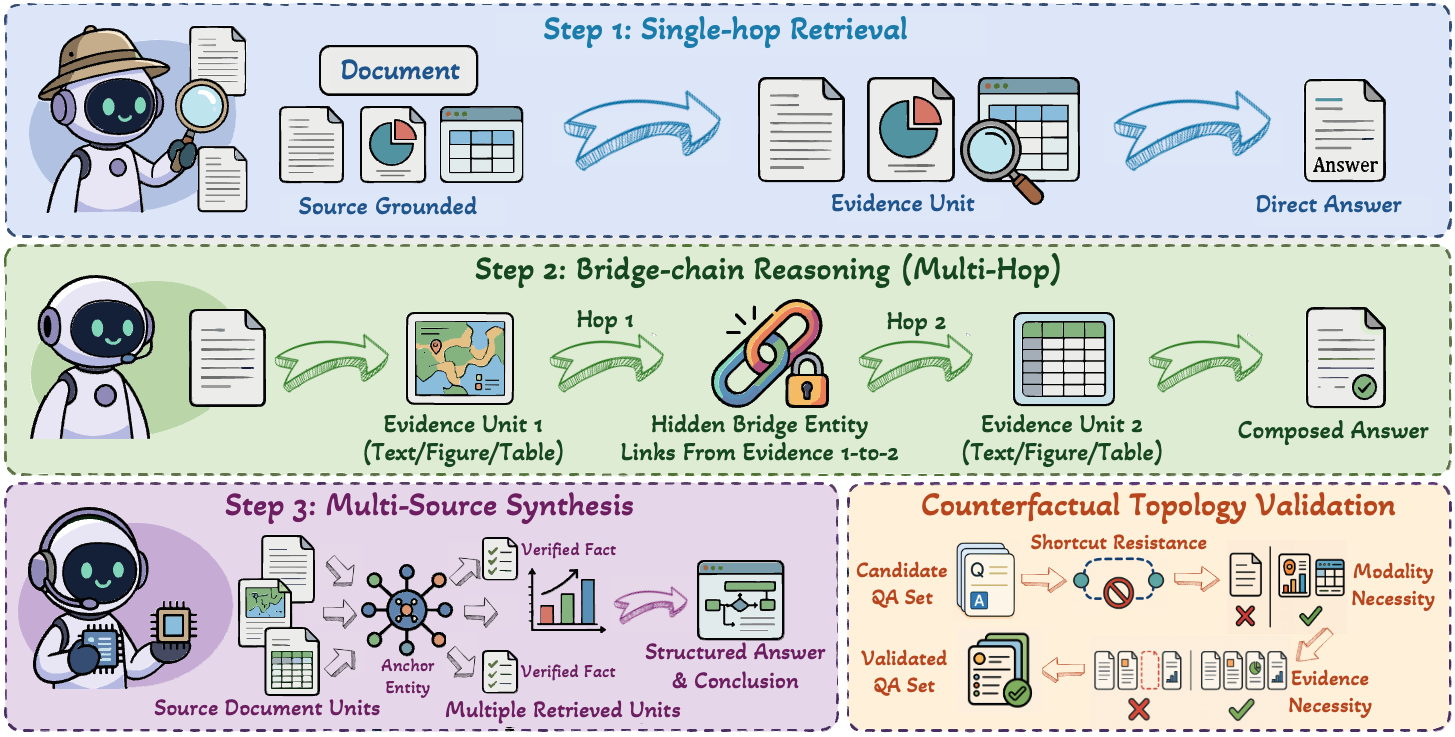}
\caption{
Overview of the TopoGraphRAG benchmark construction and validation pipeline.
Starting from layout-grounded document evidence units, the pipeline constructs
single-hop, bridge-chain, and multi-source synthesis QA instances, and then
applies counterfactual topology validation to test shortcut resistance, modality
necessity, and evidence necessity.
}
\label{fig:benchmark_pipeline}
\vspace{-0.5em}
\end{figure*}

\paragraph{Pipeline overview.}
Figure~\ref{fig:benchmark_pipeline} summarizes our benchmark construction
pipeline. Starting from layout-grounded document units, we build evidence units,
identify shared entities as cross-layout anchors, generate questions under three
controlled evidence topologies, and apply counterfactual topology validation to
remove shortcut questions, pseudo-multimodal questions, and synthesis questions
that do not require multiple evidence sources.

\paragraph{Layout-grounded evidence units.}
We build our benchmark on MMDocIR \citep{dong2025mmdocir}, a multimodal
document retrieval corpus of long, visually rich documents, and randomly sample
201 documents from the original collection. Following MMDocIR, we reinterpret
layout-level ``quotes'' as layout-grounded evidence units. Textual units retain
their original text, while visual units, including figures and tables, preserve
the original image and are augmented with OCR and VLM descriptions. Each unit
keeps its document ID, page ID, layout ID, modality label, and semantic
representation for entity extraction, evidence linking, and question
construction.

\begin{figure}[!htbp]
\centering
\includegraphics[width=0.8\linewidth]{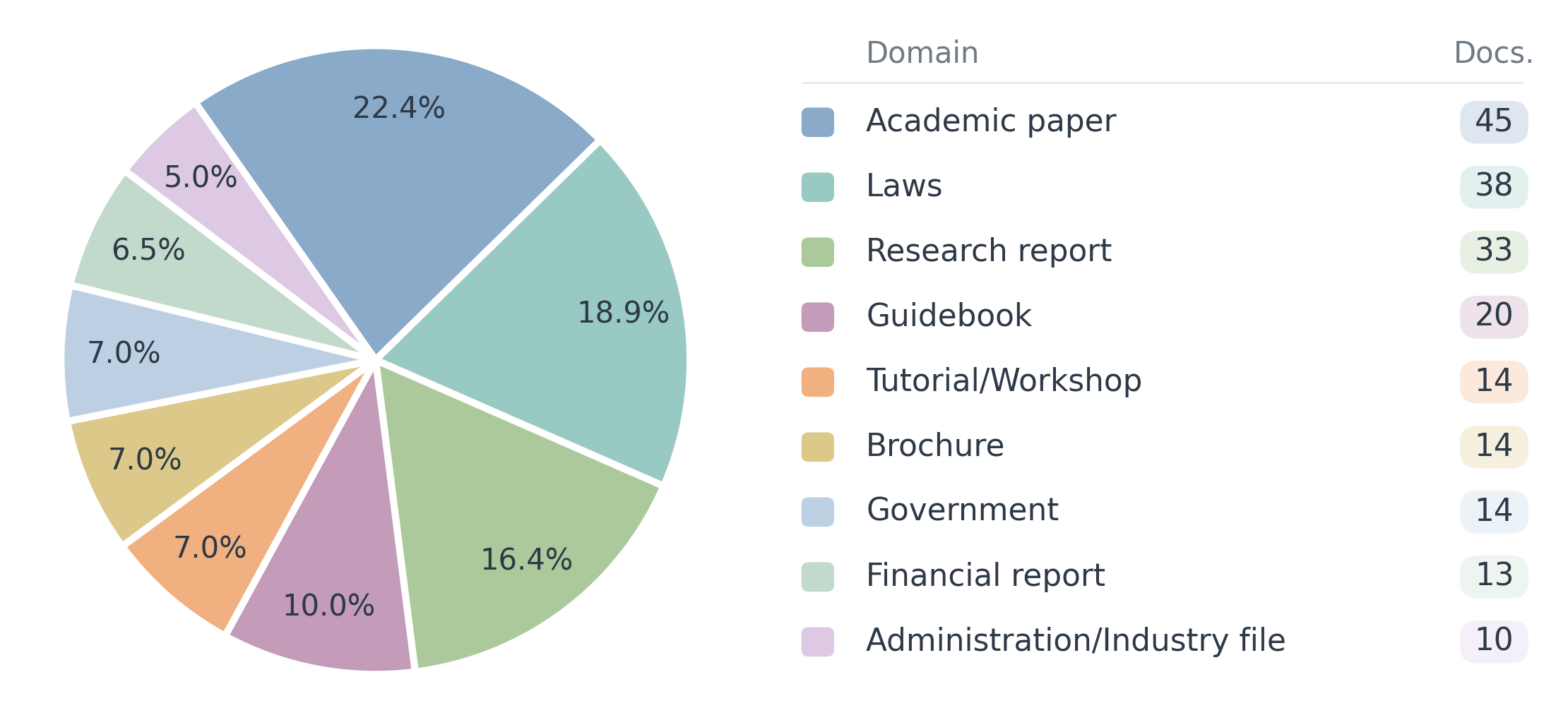}
\caption{Domain distribution of the 201 documents used to construct
\textsc{TopoGraphRAG-Bench}. Slices show the percentage of documents
in each domain, and the legend reports document counts.}
\label{fig:corpus_domains}
\end{figure}

The 201 documents cover nine of MMDocIR's ten application domains. Figure~\ref{fig:corpus_domains} shows their distribution using the original MMDocIR domain labels.

\paragraph{Anchor discovery.}
Inspired by MuSiQue's bottom-up strategy for constructing multi-hop questions
from connected single-hop questions \citep{trivedi2022musique}, we adapt this
principle to multimodal document layouts. Instead of prompting an LLM over an
entire document, we construct questions over explicit evidence topologies:
isolated units support direct retrieval questions, entity-linked units form
bridge-chain questions, and complementary evidence sets support multi-source
synthesis.

For each evidence unit, we extract salient entities and concepts using an LLM.
Textual units are processed from their original text, whereas visual units are
represented by captions, topics, OCR outputs, and VLM descriptions. The entity
schema covers people, organizations, demographic groups, locations, dates,
numerical values, metrics, and other domain-specific concepts. We normalize
entity mentions and apply fuzzy matching to identify recurring entities across
evidence units. These recurring entities serve as semantic anchors for
evidence-grounded question composition, and their page and modality distributions
allow us to construct single-page, cross-page, and cross-modal instances.

\paragraph{Single-hop retrieval construction.}
Single-hop questions are constructed from individual text, figure, or table
units to evaluate local evidence grounding without compositional dependency. The
gold evidence set is restricted to the source unit, and the answer must be
recoverable without consulting any other layout in the document.

\paragraph{Bridge-chain construction.}
Bridge-chain questions are constructed by composing two atomic QA pairs through
a shared entity. The bridge QA asks for the shared entity indirectly, without
mentioning its surface form, while the target QA explicitly mentions that entity
and asks about an attribute, value, trend, or relation grounded in another
evidence unit. An LLM then rewrites the target question by replacing the explicit
entity mention with the indirect description from the bridge QA, producing a
fluent multi-hop question with a controlled bridge dependency.

\paragraph{Multi-source synthesis construction.}
Synthesis questions are constructed to evaluate evidence aggregation beyond a
single bridge chain. Starting from shared entities, we select information-rich
entities that occur in at least five evidence units and extract concrete facts
from each occurrence; for figures and tables, the original image is additionally
provided to a VLM to improve numerical and structural fidelity. After removing
near-duplicate facts, we retain entities with at least three distinct evidence
facts and ask an LLM to identify higher-level patterns such as trends,
contradictions, causal implications, group comparisons, and summary-level
observations. Each resulting question is paired with a structured answer that
links evidence points to specific layouts and modalities, followed by a concise
synthesis conclusion.

\begin{table*}[t]
\centering
\small
\setlength{\tabcolsep}{3pt}
\renewcommand{\arraystretch}{0.98}

\begin{minipage}[t]{0.48\textwidth}
\centering
\begin{tabular}{@{}l r@{}}
\toprule
\textbf{Statistic} & \textbf{Number} \\
\midrule
\textbf{Documents} & 201 \\
- Pages/doc & 45.5 / 27 / 318 \\
- Evidence units/doc & 335.1 / 207 / 2,423 \\
\midrule
\textbf{Total Questions} & 2,024 \\
- Single-hop & 302 (14.9\%) \\
- Bridge-chain & 1,014 (50.1\%) \\
- Synthesis & 708 (35.0\%) \\
\midrule
\textbf{Evidence Distribution} & \\
- Cross-page & 1,596 (78.9\%) \\
- Cross-modal & 773 (38.2\%) \\
- Figure/table-required & 1,353 (66.8\%) \\
- Gold evidence units & 2.3 / 2 / 6 \\
- Gold pages & 2.1 / 2 / 6 \\
\midrule
\multicolumn{2}{@{}l@{}}{\textbf{Evidence Modality}} \\
Text-only & 671 (33.2\%) \\
Figure-only & 334 (16.5\%) \\
Table-only & 246 (12.2\%) \\
Cross-modal & 773 (38.2\%) \\
\bottomrule
\end{tabular}
\end{minipage}
\hfill
\begin{minipage}[t]{0.48\textwidth}
\centering
\begin{tabular}{@{}l r l r@{}}
\toprule
\multicolumn{2}{c}{\textbf{Single-hop}} &
\multicolumn{2}{c}{\textbf{Synthesis}} \\
\cmidrule(lr){1-2}\cmidrule(lr){3-4}
Modality & N & Category & N \\
\midrule
Figure-only & 122 & Text-only & 267 \\
Table-only & 113 & Cross-modal & 259 \\
Text-only & 67 & Figure-only & 113 \\
& & Table-only & 69 \\
\midrule
\multicolumn{2}{c}{\textbf{2-hop Bridge}} &
\multicolumn{2}{c}{\textbf{Synthesis Sources}} \\
\cmidrule(lr){1-2}\cmidrule(lr){3-4}
Path & N & Sources & N \\
\midrule
text$\rightarrow$text & 337 & 3 evidence units & 575 \\
text$\rightarrow$figure & 204 & 4 evidence units & 101 \\
text$\rightarrow$table & 119 & 5 evidence units & 23 \\
figure$\rightarrow$figure & 99 & 6 evidence units & 9 \\
table$\rightarrow$table & 64 & & \\
figure$\rightarrow$text & 62 & & \\
figure$\rightarrow$table & 54 & & \\
table$\rightarrow$text & 45 & & \\
table$\rightarrow$figure & 30 & & \\
\bottomrule
\end{tabular}

\vspace{0.4em}
\footnotesize
\emph{Note:} Slash-separated values denote average / median / maximum.
Bridge paths are ordered by hop sequence.
\end{minipage}

\caption{Dataset statistics and fine-grained evidence topology.}
\label{tab:dataset_stats}
\end{table*}

\subsection{Quality Assurance}
\label{Quality_Assurance}
The construction pipeline uses counterfactual validation to ensure that each
instance preserves its intended evidence topology rather than degenerating into
an easier retrieval case. Question generation, answer generation, and quality
assessment are decoupled into independent LLM calls. For bridge-chain
candidates, we verify that the bridge entity is not revealed in the composed
question, that the final answer is not identical to the bridge entity, and that
the two hops form a coherent dependency. We then test single-source
counterfactuals by providing only the first-hop or second-hop evidence; instances
that remain answerable from either source alone are removed as pseudo-multihop
cases.

We also test modality dependence and evidence necessity. For cross-modal
questions, we replace the multimodal context with text-only context and remove
cases that remain answerable without figures or tables. For synthesis questions,
we conduct leave-one-out ablations over supporting evidence points and retain
only instances with at least three required evidence points.

After automatic filtering, two PhD-level reviewers with experience in
multimodal LLMs and retrieval-augmented generation conducted a stratified
manual audit of 303 questions (15.0\% of the benchmark), including
targeted borderline cases. The audit was
stratified by reasoning topology and evidence modality. Reviewers independently
assigned binary validity judgments based on answer correctness, evidence
grounding, attribution accuracy, modality necessity, and topology consistency,
with topology-specific checks for bridge-chain and synthesis questions. The
reviewers achieved substantial agreement (Cohen's $\kappa=0.84$). Disagreements
were resolved through discussion, and audited instances were revised or removed
when their evidence or topology requirement was not valid. After automatic
validation and manual quality assurance, the strict benchmark contains 2,024
questions across 201 documents, with approximately 15\% single-hop retrieval,
50\% bridge-chain reasoning, and 35\% multi-source synthesis.

\section{Task Definition}
\label{sec:task_definition}

Given a multimodal document collection, the task is to answer a question by retrieving and reasoning over layout-grounded evidence units. Each document is represented as a set of evidence units $\mathcal{D}=\{u_i\}_{i=1}^{N}$, where each unit is associated with a modality $m_i \in \{\text{text}, \text{figure}, \text{table}\}$, a page index, a layout identifier, and a semantic representation derived from text, OCR, captions, or visual descriptions. For each question $q$, the benchmark provides a gold answer $a_q$, a set of supporting evidence units $\mathcal{E}_q$, and a reasoning type $\tau_q \in \{\text{single-hop}, \text{bridge-chain}, \text{synthesis}\}$.

Unlike standard document QA settings, where target evidence is often treated as an unordered set of relevant passages, this benchmark evaluates whether a system can recover and use the intended evidence structure. In single-hop questions, the answer is grounded in one evidence unit. In bridge-chain questions, evidence units form an ordered dependency path: an earlier hop resolves an intermediate entity or condition that determines which later evidence unit is relevant. In synthesis questions, multiple evidence units jointly support a higher-level conclusion, and no single source is sufficient to recover the full answer.

\subsection{Retrieval Task and Metrics}

The retrieval stage takes a question $q$ and the document's evidence-unit set $\mathcal{D}$ as input, and returns a ranked set of candidate contexts $\mathcal{R}_q = \{r_1, r_2, \ldots, r_k\}$. The returned contexts may be textual chunks, graph nodes, entity or relation descriptions, multimodal summaries, or page-level visual contexts, depending on the retrieval system. The retrieval objective is not merely to find evidence related to the final answer, but to retrieve the evidence units and intermediate links required by the intended evidence structure.

For single-hop questions, successful retrieval requires covering the source evidence unit. For bridge-chain questions, retrieval should cover both the bridge evidence and the target evidence, including the intermediate bridge entity that links the two hops. For synthesis questions, retrieval should cover the required evidence points that jointly support the synthesis conclusion. Thus, retrieval is evaluated as evidence-structure recovery rather than flat relevance ranking.

Retrieval quality is evaluated using RAGAS context precision and context recall \cite{es2024ragas}, instantiated with topology-enriched references. Context precision measures whether relevant contexts are ranked ahead of irrelevant ones, while context recall measures how much required reference information is covered by the retrieved contexts. To make these metrics sensitive to compositional evidence requirements, we expand the reference beyond the final answer: bridge-chain questions include reasoning traces and bridge entities, while synthesis questions include required evidence points. This enriched reference penalizes retrieval outputs that recover only final-hop evidence while missing bridge evidence or partially covering synthesis support. Formal metric definitions are
provided in Appendix~\ref{app:metrics}.

\subsection{Generation Task and Metrics}

The generation stage receives the question $q$ and the retrieved contexts $\mathcal{R}_q$, and produces an answer $\hat{a}_q$. A correct response must satisfy both answer-level and structure-level requirements: it should match the gold answer or synthesis conclusion, and it should be grounded in the retrieved evidence while covering the reasoning steps required by the question. This distinction is important for compositional document reasoning. A model may produce the correct final answer while omitting the bridge entity, skipping an intermediate hop, or relying on only one evidence source in a synthesis question. Conversely, a model may retrieve or mention relevant evidence but fail to combine it into the correct final answer. The generation task therefore evaluates answer correctness, evidence grounding, response relevance, and reasoning-step completion.

Generation quality is evaluated using answer accuracy, faithfulness, response relevancy, and step coverage. Answer accuracy measures agreement between the generated answer and the gold answer. Faithfulness measures whether generated claims are supported by the retrieved contexts, and response relevancy measures whether the response addresses the question. Step coverage evaluates whether the response covers the reasoning steps specified by the question structure. For bridge-chain questions, this includes identifying the bridge entity and conveying the corresponding target-hop result. For synthesis questions, this includes covering the required evidence points that support the intended synthesis conclusion. Formal metric definitions and the judging protocol are provided in Appendix~\ref{app:metrics}.

\section{Experiments}
\subsection{Experimental Setup}

\paragraph{Evaluated systems.}
We evaluate six retrieval-augmented systems grouped by how they represent document evidence. The text-only GraphRAG baselines, LightRAG, HippoRAG, and Microsoft GraphRAG \citep{guo2024lightrag, gutierrez2024hipporag, edge2024local}, construct graphs from textual chunks, entities, relations, or summaries, testing how far text-derived graph retrieval can support layout-grounded multimodal reasoning. VisRAG \citep{yuvisrag} represents page-level visual retrieval and tests whether direct access to document page images is sufficient for layout-rich questions. The multimodal GraphRAG systems, RAG-Anything and MegaRAG \citep{guo2025rag,hsiao2025megarag}, incorporate visual, tabular, and textual evidence into graph-structured retrieval. Together, these systems compare text-derived graphs, page-level visual retrieval, and multimodal graph structures under the same benchmark topologies.

\paragraph{Model configuration.}
We fix the model backends within each system family. Text-only GraphRAG systems use Qwen3-30B-A3B-Instruct with Qwen3-Embedding-8B. VisRAG uses Qwen3-VL-30B-A3B-Instruct for generation and VisRAG-Ret for page-image retrieval. Multimodal GraphRAG systems use Qwen3-VL-30B-A3B-Instruct with Qwen3-VL-Embedding-8B. All LLM-based evaluation metrics use Qwen3.5-35B-A3B as the judge. Detailed indexing, retrieval, and backend configurations are provided in Appendix~\ref{app:baseline_config}. Cross-judge analyses and a human spot-check of step coverage are reported in Appendix~\ref{app:judge_reliability}.

\paragraph{Evaluation organization.}
We first report aggregate retrieval and generation results, then analyze performance by reasoning topology, ordered bridge path, and synthesis composition. Additional uncertainty estimates and complete fine-grained breakdowns are provided in Appendix~\ref{app:additional_results}.

\subsection{Main Results}
\label{sec:main_results}

\begin{table*}[t]
\centering
\small
\setlength{\tabcolsep}{4pt}
\renewcommand{\arraystretch}{1.05}
\resizebox{\textwidth}{!}{%
\begin{tabular}{@{}lcccccc@{}}
\toprule
\textbf{System}
& \multicolumn{2}{c}{\textbf{Retrieval}} 
& \multicolumn{4}{c}{\textbf{Generation}} \\
\cmidrule(lr){2-3} \cmidrule(lr){4-7}
& \textbf{Context Precision} 
& \textbf{Context Recall} 
& \textbf{Answer Accuracy} 
& \textbf{Faithfulness} 
& \textbf{Response Relevancy} 
& \textbf{Step Coverage} \\
\midrule
LightRAG 
& 0.6952 & 0.6353 & 0.4045 & 0.7317 & 0.7269 & 0.6031 \\
HippoRAG 
& 0.3834 & 0.4353 & 0.3388 & 0.6517 & 0.5924 & 0.3952 \\
Microsoft GraphRAG 
& 0.6018 & 0.5839 & 0.3111 & 0.6802 & 0.6754 & 0.5402 \\
VisRAG 
& -- & -- & 0.3459 & -- & 0.7688 & 0.4337 \\
RAG-Anything 
& \textbf{0.8093} & \textbf{0.7215} & 0.5300 & 0.8346 & 0.7808 & 0.7010 \\
MegaRAG 
& 0.7125 & 0.6632 & \textbf{0.5350} & \textbf{0.9930} & \textbf{0.8172} & \textbf{0.7329} \\
\bottomrule
\end{tabular}%
}

\caption{Overall retrieval and generation performance on 2,024 questions.
Retrieval metrics are RAGAS context precision and context recall. Step coverage
is computed over bridge-chain and synthesis questions only. VisRAG is omitted
from the text-based retrieval evaluation, and its faithfulness is not reported
because its retrieved context is page-level visual input rather than textual or
structured evidence strings. Approximate 95\% confidence intervals are reported
in Appendix~\ref{app:additional_results}.}
\label{tab:overall_results}
\end{table*}

Table~\ref{tab:overall_results} reports aggregate retrieval and generation performance. Multimodal GraphRAG systems achieve the strongest overall generation results. MegaRAG obtains the highest point estimates for answer accuracy, faithfulness, response relevancy, and step coverage, while RAG-Anything achieves the best retrieval scores and comparable answer accuracy. Among text-only GraphRAG baselines, LightRAG performs best overall, but remains
below the multimodal GraphRAG systems on all generation metrics.

Retrieval and generation do not produce identical system rankings. RAG-Anything has the highest context precision and context recall, whereas MegaRAG achieves stronger faithfulness, response relevancy, and step coverage. This gap suggests that layout-grounded multimodal reasoning depends not only on retrieving relevant evidence, but also on aligning and composing the retrieved evidence according to the intended topology.

VisRAG highlights a complementary limitation of page-level visual retrieval. Although it provides direct access to page images, its answer accuracy and step coverage remain substantially below those of multimodal GraphRAG systems. This suggests that page-level visual context alone is insufficient for reliable reasoning over fine-grained layout units and cross-modal evidence dependencies. We next analyze these trends by reasoning topology, bridge path, and synthesis composition.

\subsection{Topology and Modality}
\label{sec:capability_profile}

\begin{figure*}[t]
\centering
\includegraphics[width=\textwidth]{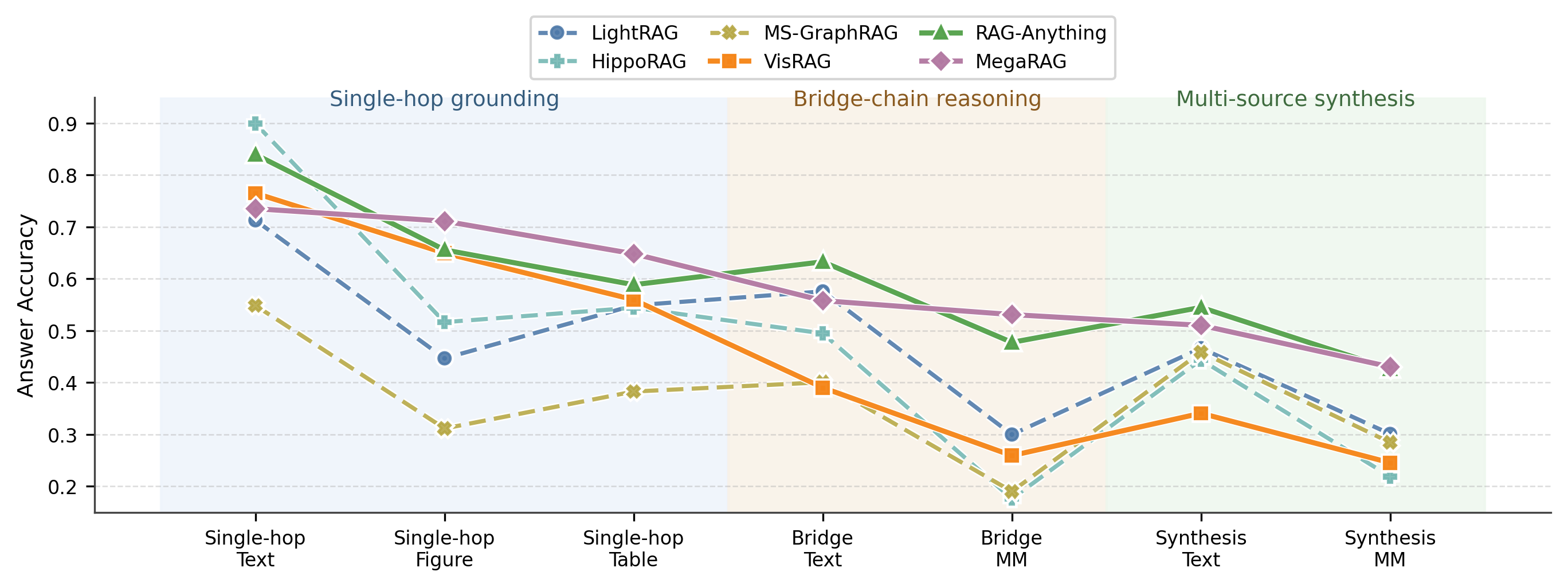}
\caption{Capability profile across reasoning topology and evidence modality.
Dashed lines denote text-only GraphRAG systems, while solid lines denote visual
or multimodal retrieval systems. Bridge-MM aggregates mixed-modality and
visual-only bridge-chain questions, and Synthesis-MM aggregates figure-only,
table-only, and cross-modal synthesis questions.}
\label{fig:capability_profile}
\end{figure*}

Figure~\ref{fig:capability_profile} shows answer accuracy across topology- and
modality-conditioned subsets. Performance generally declines as questions move
from local grounding to structured multimodal reasoning. Text-only systems can
perform strongly when the required evidence is purely textual: HippoRAG reaches
0.90 accuracy on single-hop text questions. However, the same systems degrade
sharply on multimodal bridge-chain questions. HippoRAG drops from 0.90 on
single-hop text to 0.17 on Bridge-MM, while LightRAG drops from 0.71 to 0.30.
This suggests that strong text-centered graph retrieval does not translate into
robust recovery of cross-modal evidence dependencies.

VisRAG performs competitively on single-hop visual questions, achieving 0.65 on
figure questions and 0.56 on table questions. However, its accuracy falls to
0.26 on Bridge-MM and 0.25 on Synthesis-MM, indicating that page-level visual
retrieval alone does not reliably recover fine-grained evidence topology. In
contrast, multimodal GraphRAG systems show a more stable profile on
compositional multimodal subsets. RAG-Anything reaches 0.48 on Bridge-MM and
0.43 on Synthesis-MM, while MegaRAG reaches 0.53 and 0.43, respectively.
Additional retrieval-side results by reasoning topology are reported in
Appendix~\ref{app:additional_results}. Overall, these results indicate that
multimodal document reasoning requires not only access to visual evidence, but
also a mechanism for connecting heterogeneous evidence units.

\subsection{Bridge Paths}
\label{sec:bridge_path}

\begin{figure*}[t]
\centering
\begin{minipage}[c]{0.39\textwidth}
\centering
\includegraphics[width=\linewidth]{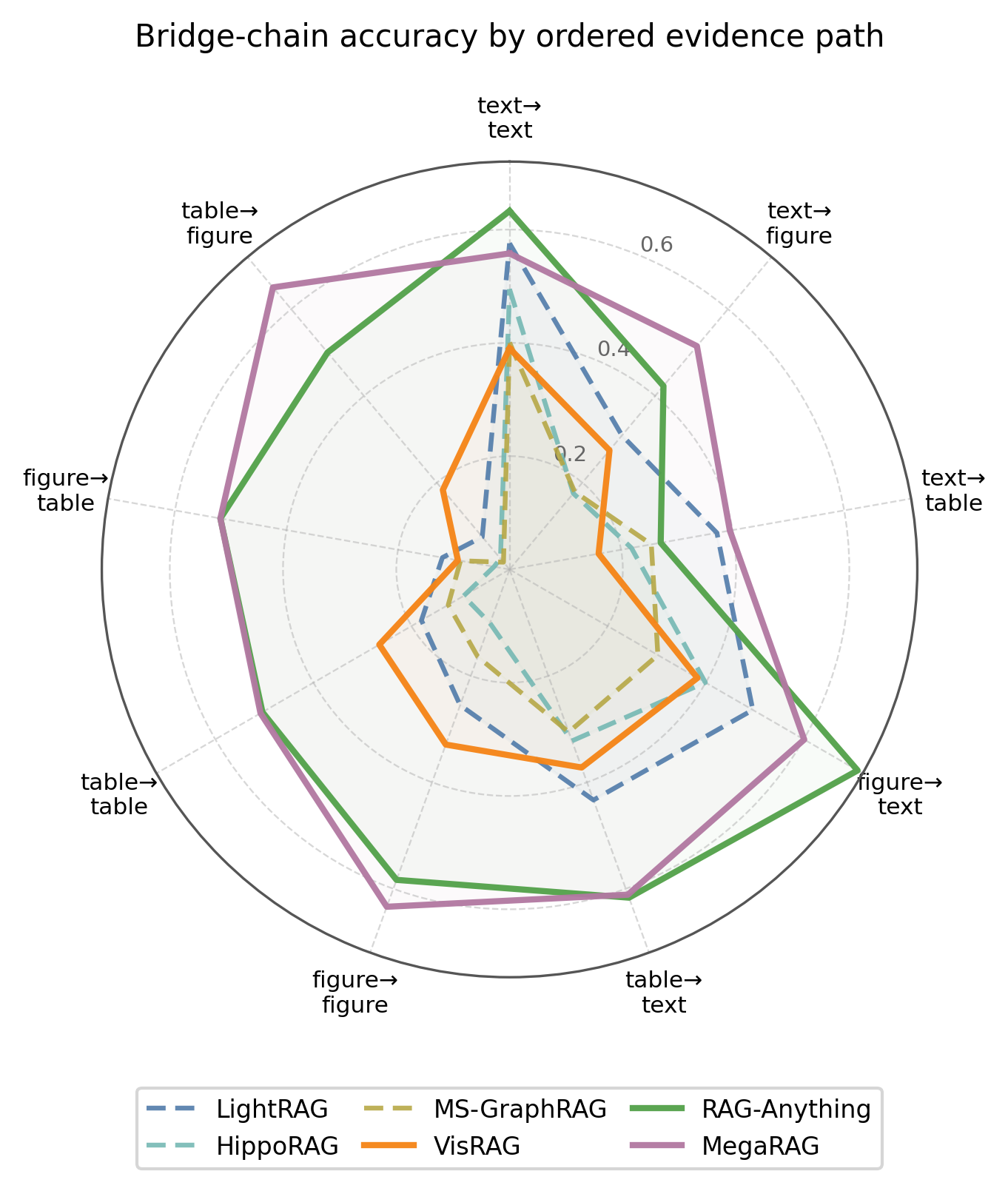}
\end{minipage}
\hfill
\begin{minipage}[c]{0.58\textwidth}
\centering
\scriptsize
\setlength{\tabcolsep}{2pt}
\renewcommand{\arraystretch}{0.88}

\textbf{Answer Accuracy (\%)}\\[-0.2em]
\resizebox{\linewidth}{!}{%
\begin{tabular}{@{}lrrrrrr@{}}
\toprule
Family & LightRAG & HippoRAG & Microsoft GraphRAG & VisRAG & RAG-Anything & MegaRAG \\
\midrule
Text-only & 57.6 & 49.5 & 40.1 & 39.0 & \textbf{63.3} & 55.8 \\
Mixed & 36.6 & 23.5 & 23.1 & 26.9 & 44.2 & \textbf{50.4} \\
Visual-only & 18.4 & 7.4 & 11.9 & 24.3 & 53.8 & \textbf{57.8} \\
\bottomrule
\end{tabular}%
}

\vspace{0.55em}
\textbf{Faithfulness (\%)}\\[-0.2em]
\resizebox{\linewidth}{!}{%
\begin{tabular}{@{}lrrrrrr@{}}
\toprule
Family & LightRAG & HippoRAG & Microsoft GraphRAG & VisRAG & RAG-Anything & MegaRAG \\
\midrule
Text-only & 77.9 & 75.2 & 75.2 & -- & 92.4 & \textbf{98.6} \\
Mixed & 74.3 & 70.3 & 72.0 & -- & 85.5 & \textbf{98.9} \\
Visual-only & 70.2 & 60.9 & 71.9 & -- & 77.7 & \textbf{99.4} \\
\bottomrule
\end{tabular}%
}

\vspace{0.55em}
\textbf{Step Coverage (\%)}\\[-0.2em]
\resizebox{\linewidth}{!}{%
\begin{tabular}{@{}lrrrrrr@{}}
\toprule
Family & LightRAG & HippoRAG & Microsoft GraphRAG & VisRAG & RAG-Anything & MegaRAG \\
\midrule
Text-only & 84.6 & 65.4 & 75.2 & 59.8 & \textbf{87.8} & 85.2 \\
Mixed & 65.1 & 44.5 & 57.2 & 49.8 & 72.6 & \textbf{79.9} \\
Visual-only & 44.9 & 23.9 & 40.1 & 49.4 & 72.1 & \textbf{83.0} \\
\bottomrule
\end{tabular}%
}
\end{minipage}

\caption{Bridge-chain reasoning across ordered evidence paths and metric families. The radar plot shows path-level answer accuracy. The tables summarize answer accuracy, faithfulness, and step coverage by bridge evidence family. Faithfulness for VisRAG is omitted because its retrieved context is represented as page-level visual input rather than textual evidence strings.}

\label{fig:bridge_path}
\end{figure*}

Figure~\ref{fig:bridge_path} analyzes bridge-chain questions from two views:
path-level answer accuracy and family-level reasoning metrics. The radar plot
shows that bridge-chain difficulty depends strongly on the ordered modality
sequence. Text$\rightarrow$text paths are relatively easier: LightRAG,
HippoRAG, RAG-Anything, and MegaRAG achieve 57.6\%, 49.5\%, 63.3\%, and
55.8\% accuracy, respectively. In contrast, paths requiring visual or tabular
bridge grounding are much harder for text-only GraphRAG systems. LightRAG drops
to 12.0\% on figure$\rightarrow$table and 7.5\% on table$\rightarrow$figure,
while HippoRAG and Microsoft GraphRAG remain below 10\% on both paths.

The summary tables show that this degradation is not limited to final-answer
accuracy. On text$\rightarrow$text paths, LightRAG obtains 84.6\% step coverage,
but drops to 44.9\% on visual-only paths. HippoRAG declines more sharply, from
65.4\% to 23.9\%. This indicates that text-only GraphRAG systems often fail to
recover or express the required hop structure when the bridge dependency is
grounded in figures or tables.

VisRAG improves access to visual page content, but does not solve bridge
dependency recovery. It reaches 32.9\% accuracy on figure$\rightarrow$figure and
26.6\% on table$\rightarrow$table, but only 9.3\% on figure$\rightarrow$table
and 18.3\% on table$\rightarrow$figure. Its step coverage also remains below
50\% on mixed and visual-only bridge families. These results suggest that
page-level visual retrieval can expose relevant visual evidence, but does not
reliably identify the specific layout unit linked by the bridge entity.

Multimodal GraphRAG systems are more robust. RAG-Anything and MegaRAG reach
51.8\% accuracy on figure$\rightarrow$table, and MegaRAG reaches 65.0\% on
table$\rightarrow$figure. At the family level, MegaRAG achieves 83.0\% step
coverage on visual-only bridge paths, compared with 44.9\% for LightRAG and
49.4\% for VisRAG. These gains indicate that multimodal graph structures help
preserve dependencies between heterogeneous evidence units. However, the
remaining variation across ordered paths shows that modality alignment remains a
central bottleneck for bridge-chain reasoning. Complete path-level accuracy and
step-coverage results are reported in Appendix~\ref{app:additional_results}.

\subsection{Multi-Source Synthesis}
\label{sec:synthesis_analysis}

Table~\ref{tab:synthesis_composition} reports synthesis accuracy by evidence
composition. Synthesis questions supported only by textual evidence are
consistently easier than those requiring visual, tabular, or cross-modal
evidence. For example, LightRAG reaches 46.6\% accuracy on text-only synthesis
but drops to 27.4\% on cross-modal synthesis, while HippoRAG drops from 44.4\%
to 18.4\%. This shows that synthesis difficulty is not only a function of
aggregating multiple facts; it also depends on whether those facts are grounded
in heterogeneous document units.

\begin{table}[H]
\centering
\small
\setlength{\tabcolsep}{4pt}
\renewcommand{\arraystretch}{0.95}
\resizebox{\columnwidth}{!}{%
\begin{tabular}{@{}lrrrrrr@{}}
\toprule
\textbf{Composition} & \textbf{LightRAG} & \textbf{HippoRAG} & \textbf{Microsoft GraphRAG} & \textbf{VisRAG} & \textbf{RAG-Anything} & \textbf{MegaRAG} \\
\midrule
Text-only & 46.6 & 44.4 & 45.9 & 34.1 & \textbf{54.5} & 51.0 \\
Figure-only & 36.3 & 29.2 & 31.9 & 30.3 & 46.2 & \textbf{46.7} \\
Table-only & 30.1 & 22.8 & 27.5 & 22.1 & 34.4 & \textbf{38.4} \\
Cross-modal & 27.4 & 18.4 & 27.2 & 22.7 & \textbf{43.2} & 42.7 \\
\bottomrule
\end{tabular}%
}
\caption{Answer accuracy on synthesis questions by evidence composition.}
\label{tab:synthesis_composition}
\end{table}

VisRAG does not close this gap despite direct access to page-level visual
inputs. It obtains 30.3\% accuracy on figure-only synthesis, 22.1\% on
table-only synthesis, and 22.7\% on cross-modal synthesis, suggesting that
page-level visual retrieval is insufficient for aggregating multiple
fine-grained evidence points. Multimodal GraphRAG systems are more robust across
evidence compositions: RAG-Anything achieves 43.2\% accuracy on cross-modal
synthesis, while MegaRAG reaches 38.4\% on table-only and 42.7\% on cross-modal
synthesis. However, both systems still trail their text-only synthesis
performance, indicating that heterogeneous evidence aggregation remains a
central challenge. A complementary breakdown by synthesis pattern is provided in Appendix~\ref{app:additional_results}.

\section{Conclusion}

In this paper, we presented a layout-grounded benchmark for evaluating multimodal evidence reasoning in GraphRAG. The benchmark contains 2,024 QA pairs across 201 long and layout-rich documents, with evidence grounded in text, figures, tables, and document layouts. It covers three controlled evidence topologies: single-hop retrieval, bridge-chain reasoning, and multi-source synthesis. Through a bottom-up construction pipeline and counterfactual validation, the benchmark is designed to test whether systems can recover and use the intended evidence structure rather than rely on shortcut retrieval or single-source evidence. Through evaluations of text-only GraphRAG, page-level visual retrieval, and multimodal GraphRAG systems, we show that current systems still face substantial challenges in layout-grounded multimodal reasoning. Multimodal GraphRAG systems achieve the strongest overall performance, but their gains are conditional on correctly grounding visual and tabular evidence, aligning evidence across modalities, and composing multiple evidence units into complete reasoning steps. Text-only GraphRAG remains limited when key dependencies are grounded outside text, while VisRAG-style page retrieval does not reliably recover fine-grained evidence topology. These results suggest that a significant gap remains between current multimodal retrieval systems and the needs of topology-aware document reasoning. We hope this benchmark will support future work on GraphRAG systems that model not only textual entities and relations, but also layout structure, modality alignment, and the reasoning roles of evidence units.

\clearpage
\bibliographystyle{plainnat}
\bibliography{references}

@inproceedings{sarthi2024raptor,
  title={Raptor: Recursive abstractive processing for tree-organized retrieval},
  author={Sarthi, Parth and Abdullah, Salman and Tuli, Aditi and Khanna, Shubh and Goldie, Anna and Manning, Christopher D},
  booktitle={The Twelfth International Conference on Learning Representations},
  year={2024}
}

@inproceedings{mathew2021docvqa,
  title={Docvqa: A dataset for vqa on document images},
  author={Mathew, Minesh and Karatzas, Dimosthenis and Jawahar, CV},
  booktitle={Proceedings of the IEEE/CVF winter conference on applications of computer vision},
  pages={2200--2209},
  year={2021}
}

@inproceedings{talmor2021multimodalqa,
  title={MultiModalQA: Complex Question Answering over Text, Tables and Images},
  author={Talmor, Alon and Yoran, Ori and Catav, Amnon and Lahav, Dan and Wang, Yizhong and Asai, Akari and Ilharco, Gabriel and Hajishirzi, Hannaneh and Berant, Jonathan},
  booktitle={International Conference on Learning Representations},
  year={2021}
}

@article{ma2024mmlongbench,
  title={Mmlongbench-doc: Benchmarking long-context document understanding with visualizations},
  author={Ma, Yubo and Zang, Yuhang and Chen, Liangyu and Chen, Meiqi and Jiao, Yizhu and Li, Xinze and Lu, Xinyuan and Liu, Ziyu and Ma, Yan and Dong, Xiaoyi and others},
  journal={Advances in Neural Information Processing Systems},
  volume={37},
  pages={95963--96010},
  year={2024}
}

@article{trivedi2022musique,
  title={MuSiQue: Multihop Questions via Single-hop Question Composition},
  author={Trivedi, Harsh and Balasubramanian, Niranjan and Khot, Tushar and Sabharwal, Ashish},
  journal={Transactions of the Association for Computational Linguistics},
  volume={10},
  pages={539--554},
  year={2022},
  doi={10.1162/tacl_a_00475}
}

@inproceedings{tangmultihop,
  title={MultiHop-RAG: Benchmarking Retrieval-Augmented Generation for Multi-Hop Queries},
  author={Tang, Yixuan and Yang, Yi},
  booktitle={First Conference on Language Modeling}
}

@article{xiang2025use,
  title={When to use graphs in rag: A comprehensive analysis for graph retrieval-augmented generation},
  author={Xiang, Zhishang and Wu, Chuanjie and Zhang, Qinggang and Chen, Shengyuan and Hong, Zijin and Huang, Xiao and Su, Jinsong},
  journal={arXiv preprint arXiv:2506.05690},
  year={2025}
}

@article{dong2025benchmarking,
  title={Benchmarking Retrieval-Augmented Multimodal Generation for Document Question Answering},
  author={Dong, Kuicai and Chang, Yujing and Huang, Shijie and Wang, Yasheng and Tang, Ruiming and Liu, Yong},
  journal={arXiv preprint arXiv:2505.16470},
  year={2025}
}

@article{peng2025unidoc,
  title={Unidoc-bench: A unified benchmark for document-centric multimodal rag},
  author={Peng, Xiangyu and Qin, Can and Chen, Zeyuan and Xu, Ran and Xiong, Caiming and Wu, Chien-Sheng},
  journal={arXiv preprint arXiv:2510.03663},
  year={2025}
}

@inproceedings{wasserman2025real,
  title={Real-mm-rag: A real-world multi-modal retrieval benchmark},
  author={Wasserman, Navve and Pony, Roi and Naparstek, Oshri and Goldfarb, Adi Raz and Schwartz, Eli and Barzelay, Udi and Karlinsky, Leonid},
  booktitle={Proceedings of the 63rd Annual Meeting of the Association for Computational Linguistics (Volume 1: Long Papers)},
  pages={31660--31683},
  year={2025}
}

@article{edge2024local,
  title={From local to global: A graph rag approach to query-focused summarization},
  author={Edge, Darren and Trinh, Ha and Cheng, Newman and Bradley, Joshua and Chao, Alex and Mody, Apurva and Truitt, Steven and Metropolitansky, Dasha and Ness, Robert Osazuwa and Larson, Jonathan},
  journal={arXiv preprint arXiv:2404.16130},
  year={2024}
}

@article{guo2024lightrag,
  title={Lightrag: Simple and fast retrieval-augmented generation},
  author={Guo, Zirui and Xia, Lianghao and Yu, Yanhua and Huang, Chao}
}

@article{gutierrez2024hipporag,
  title={Hipporag: Neurobiologically inspired long-term memory for large language models},
  author={Guti{\'e}rrez, Bernal J and Shu, Yiheng and Gu, Yu and Yasunaga, Michihiro and Su, Yu},
  journal={Advances in neural information processing systems},
  volume={37},
  pages={59532--59569},
  year={2024}
}

@article{he2024g,
  title={G-retriever: Retrieval-augmented generation for textual graph understanding and question answering},
  author={He, Xiaoxin and Tian, Yijun and Sun, Yifei and Chawla, Nitesh V and Laurent, Thomas and LeCun, Yann and Bresson, Xavier and Hooi, Bryan},
  journal={Advances in Neural Information Processing Systems},
  volume={37},
  pages={132876--132907},
  year={2024}
}

@inproceedings{luogfm,
  title={GFM-RAG: Graph Foundation Model for Retrieval Augmented Generation},
  author={Luo, Linhao and Zhao, Zicheng and Haffari, Gholamreza and Phung, Dinh and Gong, Chen and Pan, Shirui},
  booktitle={The Thirty-ninth Annual Conference on Neural Information Processing Systems}
}

@inproceedings{sunthink,
  title={Think-on-Graph: Deep and Responsible Reasoning of Large Language Model on Knowledge Graph},
  author={Sun, Jiashuo and Xu, Chengjin and Tang, Lumingyuan and Wang, Saizhuo and Lin, Chen and Gong, Yeyun and Ni, Lionel and Shum, Heung-Yeung and Guo, Jian},
  booktitle={The Twelfth International Conference on Learning Representations}
}

@inproceedings{mathew2022infographicvqa,
  title={Infographicvqa},
  author={Mathew, Minesh and Bagal, Viraj and Tito, Rub{\`e}n and Karatzas, Dimosthenis and Valveny, Ernest and Jawahar, CV},
  booktitle={Proceedings of the IEEE/CVF Winter Conference on Applications of Computer Vision},
  pages={1697--1706},
  year={2022}
}

@inproceedings{zhu2022towards,
  title={Towards complex document understanding by discrete reasoning},
  author={Zhu, Fengbin and Lei, Wenqiang and Feng, Fuli and Wang, Chao and Zhang, Haozhou and Chua, Tat-Seng},
  booktitle={Proceedings of the 30th ACM International Conference on Multimedia},
  pages={4857--4866},
  year={2022}
}

@article{tito2023hierarchical,
  title={Hierarchical multimodal transformers for multipage docvqa},
  author={Tito, Rub{\`e}n and Karatzas, Dimosthenis and Valveny, Ernest},
  journal={Pattern Recognition},
  volume={144},
  pages={109834},
  year={2023},
  publisher={Elsevier}
}

@inproceedings{zou2025docbench,
  title={Docbench: A benchmark for evaluating llm-based document reading systems},
  author={Zou, Anni and Yu, Wenhao and Zhang, Hongming and Ma, Kaixin and Cai, Deng and Zhang, Zhuosheng and Zhao, Hai and Yu, Dong},
  booktitle={Proceedings of the 4th International Workshop on Knowledge-Augmented Methods for Natural Language Processing},
  pages={359--373},
  year={2025}
}

@inproceedings{dong2025mmdocir,
  title={MMDocIR: Benchmarking Multimodal Retrieval for Long Documents},
  author={Dong, Kuicai and Chang, Yujing and Deik, Derrick Goh Xin and Li, Dexun and Tang, Ruiming and Liu, Yong},
  booktitle={Proceedings of the 2025 Conference on Empirical Methods in Natural Language Processing},
  pages={30959--30993},
  year={2025}
}

@article{loison2026vidore,
  title={ViDoRe V3: A Comprehensive Evaluation of Retrieval Augmented Generation in Complex Real-World Scenarios},
  author={Loison, Ant{\'o}nio and Mac{\'e}, Quentin and Edy, Antoine and Xing, Victor and Balough, Tom and Moreira, Gabriel and Liu, Bo and Faysse, Manuel and Hudelot, C{\'e}line and Viaud, Gautier},
  journal={arXiv preprint arXiv:2601.08620},
  year={2026}
}

@article{park2025dochop,
  title={DocHop-QA: Towards Multi-Hop Reasoning over Multimodal Document Collections},
  author={Park, Jiwon and Pyeon, Seohyun and Kim, Jinwoo and Cabal, Rina Carines and Ding, Yihao and Han, Soyeon Caren},
  journal={arXiv preprint arXiv:2508.15851},
  year={2025}
}

@article{xiang2026bridge,
  title={BRIDGE: Benchmark for multi-hop Reasoning In long multimodal Documents with Grounded Evidence},
  author={Xiang, Biao and Han, Soyeon Caren and Ding, Yihao},
  journal={arXiv preprint arXiv:2603.07931},
  year={2026}
}

@article{yang2025benchmarking,
  title={Benchmarking multimodal RAG through a chart-based document question-answering generation framework},
  author={Yang, Yuming and Zhong, Jiang and Jin, Li and Huang, Jingwang and Gao, Jingpeng and Liu, Qing and Bai, Yang and Zhang, Jingyuan and Jiang, Rui and Wei, Kaiwen},
  journal={arXiv preprint arXiv:2502.14864},
  year={2025}
}

@inproceedings{yang2018hotpotqa,
  title={HotpotQA: A dataset for diverse, explainable multi-hop question answering},
  author={Yang, Zhilin and Qi, Peng and Zhang, Saizheng and Bengio, Yoshua and Cohen, William and Salakhutdinov, Ruslan and Manning, Christopher D},
  booktitle={Proceedings of the 2018 conference on empirical methods in natural language processing},
  pages={2369--2380},
  year={2018}
}

@inproceedings{ho2020constructing,
  title={Constructing a multi-hop qa dataset for comprehensive evaluation of reasoning steps},
  author={Ho, Xanh and Nguyen, Anh-Khoa Duong and Sugawara, Saku and Aizawa, Akiko},
  booktitle={Proceedings of the 28th International Conference on Computational Linguistics},
  pages={6609--6625},
  year={2020}
}

@inproceedings{es2024ragas,
  title={Ragas: Automated evaluation of retrieval augmented generation},
  author={Es, Shahul and James, Jithin and Anke, Luis Espinosa and Schockaert, Steven},
  booktitle={Proceedings of the 18th conference of the european chapter of the association for computational linguistics: system demonstrations},
  pages={150--158},
  year={2024}
}

@article{zhou2025depth,
  title={In-depth Analysis of Graph-based RAG in a Unified Framework},
  author={Zhou, Yingli and Su, Yaodong and Sun, Youran and Wang, Shu and Wang, Taotao and He, Runyuan and Zhang, Yongwei and Liang, Sicong and Liu, Xilin and Ma, Yuchi and others},
  journal={arXiv preprint arXiv:2503.04338},
  year={2025}
}

@article{guo2025rag,
  title={Rag-anything: All-in-one rag framework},
  author={Guo, Zirui and Ren, Xubin and Xu, Lingrui and Zhang, Jiahao and Huang, Chao},
  journal={arXiv preprint arXiv:2510.12323},
  year={2025}
}

@article{hsiao2025megarag,
  title={MegaRAG: Multimodal Knowledge Graph-Based Retrieval Augmented Generation},
  author={Hsiao, Chi-Hsiang and Wang, Yi-Cheng and Lin, Tzung-Sheng and Yeh, Yi-Ren and Chen, Chu-Song},
  journal={arXiv preprint arXiv:2512.20626},
  year={2025}
}

@inproceedings{yuvisrag,
  title={VisRAG: Vision-based Retrieval-augmented Generation on Multi-modality Documents},
  author={Yu, Shi and Tang, Chaoyue and Xu, Bokai and Cui, Junbo and Ran, Junhao and Yan, Yukun and Liu, Zhenghao and Wang, Shuo and Han, Xu and Liu, Zhiyuan and others},
  booktitle={The Thirteenth International Conference on Learning Representations}
}







\appendix

\section*{Appendix Overview}

The appendix includes the following sections:
\begin{itemize}
    \item \textbf{Appendix A:} Reports baseline configuration and efficiency statistics, including indexing parameters, retrieval settings, runtime, token usage, and graph/index size.
    \item \textbf{Appendix B:} Provides additional details about evaluation metrics and topology-aware step coverage.
    \item \textbf{Appendix C:} Provides additional experimental results, including fine-grained synthesis-pattern performance.
    \item \textbf{Appendix D:} Discusses limitations of the benchmark scope, topology coverage, evaluated systems, and indexing cost.
    \item \textbf{Appendix E:} Describes license terms, data use, and the relationship between the upstream MMDocIR assets and the new annotations introduced in this work.
\end{itemize}

\section{Baseline Configuration and Efficiency}
\label{app:baseline_config}

This appendix reports the baseline configurations and efficiency statistics used
in our experiments. We use each system's official or framework-default indexing
pipeline unless a parameter is explicitly specified in our evaluation wrapper.
All systems are indexed independently at the document level. Text-only GraphRAG
systems operate on original layout-grounded text chunks, VisRAG indexes page
images, and multimodal GraphRAG systems index textual, tabular, and visual
layout units through their native multimodal pipelines.

\begin{table*}[!ht]
\centering
\scriptsize
\setlength{\tabcolsep}{3pt}
\renewcommand{\arraystretch}{1.08}
\resizebox{\textwidth}{!}{%
\begin{tabular}{@{}p{0.12\textwidth}p{0.16\textwidth}p{0.36\textwidth}p{0.17\textwidth}p{0.15\textwidth}@{}}
\toprule
\textbf{System} & \textbf{Input unit} & \textbf{Indexing / graph construction} & \textbf{Retrieval configuration} & \textbf{Model / retriever} \\
\midrule
LightRAG &
Text layout chunks &
Default LightRAG indexing pipeline for constructing a text-derived graph over chunks, entities, and relations &
Mixed retrieval mode; top-$k=3$ &
Qwen3-30B-A3B-Instruct; Qwen3-Embedding-8B \\
\midrule
HippoRAG &
Text layout chunks &
Default HippoRAG indexing pipeline over text chunks and its native retrieval graph &
Default retrieval; top-$k=3$ &
Qwen3-30B-A3B-Instruct; Qwen3-Embedding-8B \\
\midrule
MS-GraphRAG &
Concatenated text file &
Default Microsoft GraphRAG indexing workflow for text-unit chunking, entity and relationship extraction, and community construction &
Local search; top-3 entities and top-3 relationships; max context tokens 12000 &
Qwen3-30B-A3B-Instruct; Qwen3-Embedding-8B \\
\midrule
VisRAG &
Page images &
Default VisRAG page-image indexing using VisRAG-Ret embeddings; no entity-relation graph is constructed. &
Cosine page retrieval; top-$k=3$ pages. &
Qwen3-VL-30B-A3B-Instruct; VisRAG-Ret \\
\midrule
RAG-Anything &
Multimodal layout chunks &
Default RAG-Anything multimodal indexing pipeline, with LightRAG as the internal graph backbone and native processing for text, figures, and tables &
Mixed retrieval mode; top-$k=3$ &
Qwen3-VL-30B-A3B-Instruct; Qwen3-VL-Embedding-8B \\
\midrule
MegaRAG &
Page-content JSON &
Default MegaRAG multimodal knowledge-graph construction over page-level text, figure, table, and image representations &
Mixed retrieval mode; chunk top-$k=6$; reranking disabled &
Qwen3-VL-30B-A3B-Instruct; Qwen3-VL-Embedding-8B \\
\bottomrule
\end{tabular}%
}
\caption{Indexing, graph construction, and retrieval configurations for evaluated systems. We use each system's default or native indexing and graph-construction pipeline without algorithmic modification, and standardize only the document input, model backend, and retrieval settings needed for fair comparison.}

\label{tab:baseline_config}
\end{table*}

Table~\ref{tab:rag_efficiency_combined} reports the efficiency and graph-size
statistics of the evaluated systems. We separate indexing time from retrieval
and generation time because GraphRAG systems often move substantial computation
to the offline indexing stage, where entities, relationships, communities, or
multimodal graph structures are constructed. Runtime, token usage, and graph
statistics are reported per document using the 201-document benchmark
denominator. Missing entries indicate that the quantity is either not
applicable to the system design or not recorded in the available artifact.

\begin{table*}[t]
\centering
\scriptsize
\setlength{\tabcolsep}{3pt}
\renewcommand{\arraystretch}{1.08}
\resizebox{\textwidth}{!}{%
\begin{tabular}{@{}lrrr rrr rrrr@{}}
\toprule
& \multicolumn{3}{c}{\textbf{Runtime (s/doc)}} &
\multicolumn{3}{c}{\textbf{LLM Usage (/doc)}} &
\multicolumn{4}{c}{\textbf{Graph / Index Size (/doc)}} \\
\cmidrule(lr){2-4}\cmidrule(lr){5-7}\cmidrule(lr){8-11}
\textbf{System} &
\textbf{Indexing} & \textbf{Retrieval} & \textbf{Generation} &
\textbf{Calls} & \textbf{Prompt Tok.} & \textbf{Completion Tok.} &
\textbf{Entities} & \textbf{Relationships} & \textbf{Chunks} & \textbf{Communities} \\
\midrule
LightRAG &
2347.73 & 3.81 & 8.94 &
206.43 & 848415.46 & 100313.98 &
847.55 & 865.26 & 116.55 & -- \\
HippoRAG &
117.95 & 2.47 & 3.85 &
552.51 & 255395.94 & 56394.44 &
1686.16 & 2108.58 & 266.32 & -- \\
MS-GraphRAG &
1942.94 & 50.82 & 11.90 &
531.37 & 909167.87 & 412309.60 &
667.77 & 1004.43 & 94.24 & 115.56 \\
RAG-Anything &
5844.99 & 18.74 & 21.70 &
668.80 & 2470006.99 & 286043.06 &
1542.71 & 2786.12 & 287.04 & -- \\
VisRAG &
2.94 & 0.15 & 22.87 &
-- & -- & -- &
-- & -- & -- & -- \\
MegaRAG &
3241.78 & 335.84 & 33.57 &
15.20 & 230093.20 & 5026.03 &
804.59 & 1142.22 & 45.55 & -- \\
\bottomrule
\end{tabular}%
}
\caption{Efficiency and index statistics for evaluated RAG systems. Runtime, token usage, and graph/index sizes are normalized by the 201-document benchmark denominator. ``--'' denotes not applicable or not recorded in the available artifact. VisRAG is a visual page retrieval baseline and does not construct an entity-relation graph.}
\label{tab:rag_efficiency_combined}
\end{table*}

The efficiency profile highlights different cost-performance tradeoffs across
systems. Text-only GraphRAG systems construct entity-relation graphs from
textual chunks, while multimodal GraphRAG systems incur additional indexing and
token costs from processing visual and tabular evidence. RAG-Anything produces
the largest graph and highest token usage, reflecting its multimodal extension
over a LightRAG-style backbone. MegaRAG constructs a smaller multimodal graph but
has higher retrieval cost due to its multimodal graph and page-level retrieval
process. VisRAG has low indexing and retrieval cost because it indexes page
images without constructing an entity-relation graph, but this also means it does
not expose fine-grained graph statistics.

\section{Evaluation Metrics and Topology-Aware Judging}
\label{app:metrics}

\subsection{Metric Definitions and Judging Protocol}
\label{app:metric_definitions}

For each question $q$, the benchmark provides a gold answer $a_q$, a set of
gold evidence units $\mathcal{E}_q$, and a reasoning topology
$\tau_q \in \{\text{single}, \text{bridge}, \text{synthesis}\}$. A system
returns an ordered list of retrieved contexts
$\mathcal{R}_q=(r_1,\ldots,r_K)$ and a generated answer $\hat{a}_q$.

\paragraph{Topology-enriched reference.}
Standard RAG evaluation usually compares retrieved contexts against the final
answer alone. In our benchmark, the reference also includes the reasoning
structure required by the question. We therefore define a topology-enriched
reference $y_q$ as
\[
y_q =
\begin{cases}
(a_q, \mathcal{E}_q), & \tau_q=\text{single},\\
(a_q, \mathcal{E}_q, b_q, \pi_q), & \tau_q=\text{bridge},\\
(a_q, \mathcal{E}_q, P_q), & \tau_q=\text{synthesis},
\end{cases}
\]
where $b_q$ is the bridge entity, $\pi_q$ is the ordered bridge reasoning trace,
and $P_q$ is the set of required synthesis evidence points. This reference is
used for retrieval evaluation and for topology-aware judging.

\paragraph{Context precision.}
Following the RAGAS context precision metric, we evaluate whether relevant
contexts are ranked before irrelevant ones. Let
$v_k=\mathrm{Rel}(r_k,q,y_q)\in\{0,1\}$ indicate whether retrieved context
$r_k$ is relevant to the topology-enriched reference. Precision at rank $k$ is
\[
P_k=\frac{1}{k}\sum_{i=1}^{k} v_i .
\]
The context precision score is
\[
\mathrm{CP}(q)=
\begin{cases}
\frac{\sum_{k=1}^{K} P_k \cdot v_k}{\sum_{k=1}^{K} v_k}, &
\sum_{k=1}^{K} v_k > 0,\\
0, & \text{otherwise}.
\end{cases}
\]

\paragraph{Context recall.}
Following RAGAS context recall, we decompose the topology-enriched reference
$y_q$ into atomic reference claims $\mathcal{C}_q$. Let
$u_c=\mathrm{Supp}(c,\mathcal{R}_q)\in\{0,1\}$ indicate whether claim $c$ is
supported by the retrieved contexts. Context recall is
\[
\mathrm{CR}(q)=\frac{1}{|\mathcal{C}_q|}
\sum_{c\in \mathcal{C}_q} u_c .
\]
For bridge-chain questions, $\mathcal{C}_q$ includes claims about both the
bridge evidence and the target evidence. For synthesis questions,
$\mathcal{C}_q$ includes the required evidence points and the synthesis
conclusion.

\paragraph{Answer accuracy.}
Answer accuracy measures agreement between $\hat{a}_q$ and $a_q$. We use a
normalized LLM-as-judge score
\[
\mathrm{Acc}(q)=J_{\mathrm{acc}}(q,\hat{a}_q,a_q)\in[0,1],
\]
where the judge assigns full credit to semantically equivalent answers, zero
credit to incorrect or irrelevant answers, and partial credit when a response
captures only part of a multi-evidence answer. Direct single-hop answers are
judged primarily by semantic equivalence to the gold answer, while bridge and
synthesis answers are judged against the full gold conclusion.

\paragraph{Faithfulness.}
Following the RAGAS faithfulness metric, we decompose the generated answer
$\hat{a}_q$ into atomic response claims $\mathcal{D}_q$. Let
$w_d=\mathrm{Supp}(d,\mathcal{R}_q)\in\{0,1\}$ indicate whether response claim
$d$ is supported by the retrieved context. Faithfulness is
\[
\mathrm{Faith}(q)=\frac{1}{|\mathcal{D}_q|}
\sum_{d\in \mathcal{D}_q} w_d .
\]
This metric measures grounding in the retrieved evidence rather than agreement
with the gold answer.

\paragraph{Response relevancy.}
Following the RAGAS response relevancy metric, the judge generates
$N$ question variants $\{g_i(\hat{a}_q)\}_{i=1}^{N}$ from the response and
compares them with the original question by embedding similarity. Let $E_o$ be
the embedding of the original question and $E_{g_i}$ be the embedding of the
$i$-th generated question. The response relevancy score is
\[
\mathrm{RelAns}(q)=\frac{1}{N}\sum_{i=1}^{N}
\frac{E_{g_i}\cdot E_o}{\|E_{g_i}\|\,\|E_o\|}.
\]
Intuitively, a response is relevant when the original question can be
reconstructed from the answer content.


\paragraph{Topology-aware step coverage.}
RAGAS metrics evaluate retrieval quality, grounding, and answer relevance, but they do not directly score whether the response covers the reasoning steps specified by the question structure. We therefore define a required step set $\mathcal{S}_q$:
\[
\mathcal{S}_q =
\begin{cases}
\{s_{\mathrm{bridge}}, s_{\mathrm{target}}\}, & \tau_q=\text{bridge},\\
\{s_p:p\in P_q\}, & \tau_q=\text{synthesis}.
\end{cases}
\]
For bridge-chain questions, the required steps correspond to the bridge and target hop results. For synthesis questions, they correspond to the required evidence points; the synthesis conclusion is evaluated by answer accuracy rather than counted as an additional step. Because single-hop questions do not contain an intermediate or multi-source reasoning structure, step coverage is reported only for bridge-chain and synthesis questions.

For each bridge-chain question, the judge receives four inputs: the question, the gold answer, the annotated per-hop reasoning trace, and the system response. Each hop in the trace specifies a sub-question, its gold result, and the source modality. The judge compares the response with the annotated trace and marks a hop as covered when the response clearly conveys the corresponding gold result, allowing for differences in wording. Judgments are based solely on these four inputs. Synthesis questions follow the same protocol, with the annotated evidence points replacing the per-hop reasoning trace and a covered/not-covered decision assigned to each required evidence point.

Let $z_s=1$ if the judge marks step or evidence point $s$ as covered, and $z_s=0$ otherwise. Coverage does not require the response to present an explicitly enumerated reasoning chain. Step coverage is
\[
\mathrm{SC}(q)=\frac{1}{|\mathcal{S}_q|}
\sum_{s\in \mathcal{S}_q} z_s .
\]
Answer accuracy and step coverage jointly assess final-answer correctness and coverage of the required reasoning content. A response may provide the correct final answer while omitting intermediate results, or cover some required steps without reaching the correct answer.

\paragraph{Metric applicability.}
Text-based context precision, context recall, and faithfulness require retrieved
contexts that can be represented as textual or structured evidence strings.
VisRAG retrieves page-level visual inputs rather than fine-grained textual
evidence units, so its retrieval RAGAS scores and text-based faithfulness are
not directly comparable to the GraphRAG systems and are not reported.

\subsection{Reliability of Step Coverage Evaluation}
\label{app:judge_reliability}

We assess the reliability of step coverage using three judges from distinct model families: Qwen3.5-35B-A3B, GLM-5.1, and GPT-5.1. All six systems' responses are held fixed, and each judge applies the scoring protocol defined above at temperature 0. The analysis covers 1,722 multi-step questions, comprising 1,014 bridge-chain and 708 synthesis questions. The remaining 302 single-hop questions are excluded.

\paragraph{Sensitivity to judge choice.}
Table~\ref{tab:judge_sensitivity} reports mean step coverage under each judge. Although absolute scores vary, MegaRAG and RAG-Anything consistently achieve the highest and second-highest scores, respectively. Their advantage over the text-only GraphRAG and page-level visual retrieval baselines therefore persists across evaluator families.

\begin{table}[H]
\centering
\small
\caption{Mean step coverage under three model judges, evaluated on fixed system responses. Qwen3.5 denotes Qwen3.5-35B-A3B.}
\label{tab:judge_sensitivity}
\begin{tabular}{lccc}
\toprule
System & Qwen3.5 & GLM-5.1 & GPT-5.1 \\
\midrule
LightRAG           & 0.6031 & 0.6121 & 0.6743 \\
HippoRAG           & 0.3952 & 0.4153 & 0.4858 \\
Microsoft GraphRAG & 0.5402 & 0.5721 & 0.6029 \\
VisRAG             & 0.4337 & 0.5024 & 0.4783 \\
RAG-Anything       & 0.7010 & 0.7349 & 0.7442 \\
MegaRAG            & 0.7329 & 0.7723 & 0.7601 \\
\bottomrule
\end{tabular}
\end{table}

\paragraph{Hop-level cross-judge agreement.}
We additionally examine agreement on individual binary decisions. Each evaluation unit consists of one annotated bridge-chain hop or synthesis evidence point within a fixed system response. For each judge pair, we report the proportion of matching covered/not-covered decisions and Cohen's $\kappa$, both overall and separately by reasoning topology.

Table~\ref{tab:hop_judge_agreement} shows overall agreement ranging from 0.8512 to 0.8958, with $\kappa$ ranging from 0.7054 to 0.7923. Agreement remains high under both topologies: topology-specific agreement ranges from 0.8479 to 0.9028, and $\kappa$ ranges from 0.7002 to 0.8039. These results indicate consistency at the level of individual reasoning steps, complementing the system-level comparison above.

\begin{table}[H]
\centering
\small
\setlength{\tabcolsep}{5pt}
\caption{Pairwise hop-level agreement and Cohen's $\kappa$, macro-averaged with equal weight across the six systems. Results are reported overall and by reasoning topology.}
\label{tab:hop_judge_agreement}
\begin{tabular}{lcccccc}
\toprule
& \multicolumn{2}{c}{Overall}
& \multicolumn{2}{c}{Bridge-chain}
& \multicolumn{2}{c}{Synthesis} \\
\cmidrule(lr){2-3}
\cmidrule(lr){4-5}
\cmidrule(lr){6-7}
Judge pair
& Agreement & $\kappa$
& Agreement & $\kappa$
& Agreement & $\kappa$ \\
\midrule
Qwen3.5 / GLM-5.1
& 0.8592 & 0.7208
& 0.8663 & 0.7328
& 0.8529 & 0.7105 \\
Qwen3.5 / GPT-5.1
& 0.8512 & 0.7054
& 0.8479 & 0.7002
& 0.8541 & 0.7103 \\
GLM-5.1 / GPT-5.1
& 0.8958 & 0.7923
& 0.9028 & 0.8039
& 0.8896 & 0.7814 \\
\bottomrule
\end{tabular}
\end{table}

\paragraph{Human spot-check.}
We manually annotated a stratified sample of 100 evaluation units, consisting of 50 bridge-chain hops and 50 synthesis evidence points. The sample covered all six systems and included cases where the model judges agreed and cases where they disagreed. Each unit received a binary covered/not-covered label using the same criterion applied by the model judges.

Table~\ref{tab:human_judge_agreement} compares the human labels with each judge's decisions on the same sample. Agreement ranges from 0.83 to 0.91, with Cohen's $\kappa$ ranging from 0.6561 to 0.8190. These results provide complementary evidence of alignment between model-based step-coverage judgments and human annotation.

\begin{table}[H]
\centering
\small
\caption{Agreement between human annotations and model judges on 100 evaluation units: 50 bridge-chain hops and 50 synthesis evidence points.}
\label{tab:human_judge_agreement}
\begin{tabular}{lcc}
\toprule
Comparison & Agreement & Cohen's $\kappa$ \\
\midrule
Human / Qwen3.5 & 0.8300 & 0.6561 \\
Human / GLM-5.1 & 0.8800 & 0.7580 \\
Human / GPT-5.1 & 0.9100 & 0.8190 \\
\bottomrule
\end{tabular}
\end{table}

\section{Additional Experimental Results}
\label{app:additional_results}

This appendix provides additional fine-grained experimental results that
complement the main analysis. These results are not required for the main
comparison across system families, but they help characterize where different
reasoning patterns remain challenging.

\subsection{Overall Metric Uncertainty}

Table~\ref{tab:overall_metric_ci_app} reports approximate 95\% confidence
intervals for the overall metrics in Table~\ref{tab:overall_results}. Intervals
are computed as mean $\pm$ 1.96 standard errors over non-missing per-question
scores. Step coverage is defined for bridge-chain and synthesis questions; the
single-hop subset is excluded from this metric, so the overall step-coverage
intervals use the 1,722 non-single-hop questions.

\begin{table}[H]
\centering
\small
\setlength{\tabcolsep}{3pt}
\renewcommand{\arraystretch}{0.95}
\resizebox{\textwidth}{!}{%
\begin{tabular}{@{}lcccccc@{}}
\toprule
\textbf{System} &
\textbf{Context Precision} &
\textbf{Context Recall} &
\textbf{Answer Accuracy} &
\textbf{Faithfulness} &
\textbf{Response Relevancy} &
\textbf{Step Coverage} \\
\midrule
LightRAG & 0.695 [0.675, 0.715] & 0.635 [0.619, 0.652] & 0.405 [0.389, 0.420] & 0.732 [0.720, 0.744] & 0.727 [0.715, 0.739] & 0.603 [0.585, 0.621] \\
HippoRAG & 0.383 [0.362, 0.405] & 0.435 [0.419, 0.452] & 0.339 [0.322, 0.356] & 0.652 [0.637, 0.666] & 0.592 [0.576, 0.609] & 0.395 [0.377, 0.413] \\
Microsoft GraphRAG & 0.602 [0.580, 0.623] & 0.584 [0.567, 0.601] & 0.311 [0.298, 0.324] & 0.680 [0.669, 0.691] & 0.675 [0.662, 0.689] & 0.540 [0.522, 0.558] \\
VisRAG & -- & -- & 0.346 [0.330, 0.362] & -- & 0.769 [0.759, 0.779] & 0.434 [0.416, 0.451] \\
RAG-Anything & \textbf{0.809} [0.792, 0.826] & \textbf{0.722} [0.707, 0.736] & 0.530 [0.515, 0.545] & 0.835 [0.826, 0.843] & 0.781 [0.772, 0.790] & 0.701 [0.685, 0.717] \\
MegaRAG & 0.713 [0.693, 0.732] & 0.663 [0.647, 0.679] & \textbf{0.535} [0.521, 0.549] & \textbf{0.993} [0.991, 0.995] & \textbf{0.817} [0.811, 0.823] & \textbf{0.733} [0.718, 0.748] \\
\bottomrule
\end{tabular}%
}
\caption{Approximate 95\% confidence intervals for the overall metrics in
Table~\ref{tab:overall_results}. Step coverage is
computed over bridge-chain and synthesis questions only.}
\label{tab:overall_metric_ci_app}
\end{table}

\subsection{Retrieval Performance by Reasoning Topology}

Table~\ref{tab:retrieval_by_topology_app} reports retrieval performance by
reasoning topology. RAG-Anything obtains the strongest retrieval scores on
bridge-chain and synthesis questions, while MegaRAG is strongest on single-hop
questions. The contrast with the generation results in
Table~\ref{tab:overall_results} suggests that retrieval coverage alone does not
determine final answer quality; systems must also compose the retrieved evidence
according to the intended topology.

\begin{table}[H]
\centering
\small
\setlength{\tabcolsep}{4pt}
\renewcommand{\arraystretch}{0.98}
\resizebox{\textwidth}{!}{%
\begin{tabular}{@{}lcccccc@{}}
\toprule
\textbf{System} &
\multicolumn{2}{c}{\textbf{Single-hop}} &
\multicolumn{2}{c}{\textbf{Bridge-chain}} &
\multicolumn{2}{c}{\textbf{Synthesis}} \\
\cmidrule(lr){2-3}\cmidrule(lr){4-5}\cmidrule(lr){6-7}
& \textbf{Context Precision} & \textbf{Context Recall}
& \textbf{Context Precision} & \textbf{Context Recall}
& \textbf{Context Precision} & \textbf{Context Recall} \\
\midrule
LightRAG & 0.765 & 0.724 & 0.749 & 0.700 & 0.589 & 0.505 \\
HippoRAG & 0.702 & 0.711 & 0.449 & 0.500 & 0.154 & 0.226 \\
Microsoft GraphRAG & 0.729 & 0.707 & 0.618 & 0.617 & 0.524 & 0.484 \\
RAG-Anything & 0.848 & 0.814 & \textbf{0.881} & \textbf{0.793} & \textbf{0.691} & \textbf{0.580} \\
MegaRAG & \textbf{0.911} & \textbf{0.906} & 0.740 & 0.768 & 0.589 & 0.410 \\
\bottomrule
\end{tabular}%
}
\caption{Retrieval performance by reasoning topology. VisRAG is omitted because
its retrieved context is page-level visual input rather than textual or
structured evidence strings.}
\label{tab:retrieval_by_topology_app}
\end{table}

\subsection{Complete Bridge-Path Results}

Table~\ref{tab:bridge_path_accuracy_app} reports answer accuracy for all
ordered bridge paths. The full breakdown shows that visual and tabular paths are
not uniformly difficult: paths such as figure$\rightarrow$text and
table$\rightarrow$text are easier than figure$\rightarrow$table or
table$\rightarrow$figure, where both hops require non-textual grounding or
cross-modal alignment.

\begin{table}[H]
\centering
\small
\setlength{\tabcolsep}{4pt}
\renewcommand{\arraystretch}{0.95}
\resizebox{\textwidth}{!}{%
\begin{tabular}{@{}lrrrrrrr@{}}
\toprule
\textbf{Bridge Path} & \textbf{Number} & \textbf{LightRAG} & \textbf{HippoRAG} & \textbf{Microsoft GraphRAG} & \textbf{VisRAG} & \textbf{RAG-Anything} & \textbf{MegaRAG} \\
\midrule
text$\rightarrow$text & 337 & 57.6 & 49.5 & 40.1 & 39.0 & \textbf{63.3} & 55.8 \\
text$\rightarrow$figure & 204 & 30.9 & 17.5 & 17.9 & 27.5 & 42.3 & \textbf{51.5} \\
text$\rightarrow$table & 119 & 37.2 & 21.9 & 25.4 & 16.0 & 27.1 & \textbf{39.5} \\
figure$\rightarrow$text & 62 & 49.6 & 39.9 & 30.2 & 38.3 & \textbf{71.0} & 60.1 \\
figure$\rightarrow$figure & 99 & 25.5 & 10.1 & 16.4 & 32.9 & 58.3 & \textbf{63.4} \\
figure$\rightarrow$table & 54 & 12.0 & 2.8 & 8.8 & 9.3 & \textbf{51.8} & \textbf{51.8} \\
table$\rightarrow$text & 45 & 43.3 & 32.2 & 30.6 & 37.2 & \textbf{61.7} & 61.1 \\
table$\rightarrow$figure & 30 & 7.5 & 2.5 & 1.7 & 18.3 & 50.0 & \textbf{65.0} \\
table$\rightarrow$table & 64 & 18.0 & 9.4 & 12.5 & 26.6 & 50.4 & \textbf{50.8} \\
\bottomrule
\end{tabular}%
}
\caption{Answer accuracy (\%) on bridge-chain questions by ordered evidence
path.}
\label{tab:bridge_path_accuracy_app}
\end{table}

Table~\ref{tab:bridge_path_step_app} provides the corresponding step-coverage
breakdown. Compared with final-answer accuracy, step coverage more directly
reflects whether the model recovers both bridge and target evidence roles.

\begin{table}[H]
\centering
\small
\setlength{\tabcolsep}{4pt}
\renewcommand{\arraystretch}{0.95}
\resizebox{\textwidth}{!}{%
\begin{tabular}{@{}lrrrrrrr@{}}
\toprule
\textbf{Bridge Path} & \textbf{Number} & \textbf{LightRAG} & \textbf{HippoRAG} & \textbf{Microsoft GraphRAG} & \textbf{VisRAG} & \textbf{RAG-Anything} & \textbf{MegaRAG} \\
\midrule
text$\rightarrow$text & 337 & 84.6 & 65.4 & 75.2 & 59.8 & \textbf{87.8} & 85.2 \\
text$\rightarrow$figure & 204 & 62.5 & 40.0 & 53.4 & 50.5 & 68.4 & \textbf{80.6} \\
text$\rightarrow$table & 119 & 62.6 & 44.5 & 58.4 & 40.8 & 64.7 & \textbf{71.4} \\
figure$\rightarrow$text & 62 & 71.0 & 54.8 & 64.5 & 58.1 & \textbf{91.1} & 88.7 \\
figure$\rightarrow$figure & 99 & 51.0 & 25.2 & 40.4 & 60.1 & 76.8 & \textbf{87.9} \\
figure$\rightarrow$table & 54 & 35.2 & 10.2 & 35.2 & 29.6 & 68.5 & \textbf{80.6} \\
table$\rightarrow$text & 45 & 75.6 & 51.1 & 61.1 & 58.9 & \textbf{86.7} & \textbf{86.7} \\
table$\rightarrow$figure & 30 & 43.3 & 26.7 & 35.0 & 48.3 & 73.3 & \textbf{83.3} \\
table$\rightarrow$table & 64 & 44.5 & 32.0 & 46.1 & 50.0 & 67.2 & \textbf{77.3} \\
\bottomrule
\end{tabular}%
}
\caption{Step coverage (\%) on bridge-chain questions by ordered evidence
path.}
\label{tab:bridge_path_step_app}
\end{table}

\subsection{Synthesis Pattern Breakdown}

Table~\ref{tab:synthesis_pattern_app} reports answer accuracy on synthesis
questions by synthesis pattern. The results show that synthesis difficulty
varies not only by evidence modality, but also by the type of relation that must
be preserved across sources. Comparison and contradiction-style questions are
often challenging because they require the model to maintain distinctions among
multiple evidence points, whereas trend and causal questions can be easier when
the relevant pattern is stated more explicitly in the evidence.

\begin{table}[H]
\centering
\small
\setlength{\tabcolsep}{4pt}
\renewcommand{\arraystretch}{0.95}
\resizebox{\columnwidth}{!}{%
\begin{tabular}{@{}lrrrrrrr@{}}
\toprule
\textbf{Pattern} & \textbf{Number} & \textbf{LightRAG} & \textbf{HippoRAG} & \textbf{Microsoft GraphRAG} & \textbf{VisRAG} & \textbf{RAG-Anything} & \textbf{MegaRAG} \\
\midrule
Panorama & 354 & 37.6 & 29.1 & 36.7 & 29.7 & 48.5 & \textbf{49.6} \\
Comparison & 130 & 28.7 & 19.8 & 27.5 & 18.3 & \textbf{39.2} & 34.8 \\
Contradiction & 88 & 34.7 & 34.7 & 29.8 & 28.1 & \textbf{45.2} & 40.3 \\
Trend & 73 & 44.5 & 38.7 & 41.4 & 32.2 & \textbf{50.7} & 48.6 \\
Causal & 63 & 37.7 & 43.7 & 41.3 & 34.9 & \textbf{54.0} & \textbf{54.0} \\
\bottomrule
\end{tabular}%
}
\caption{Answer accuracy (\%) on synthesis questions by synthesis pattern.}
\label{tab:synthesis_pattern_app}
\end{table}

\subsection{Cross-Corpus Generalization}
\label{app:cross_corpus}

To assess whether our construction methodology and main empirical findings extend beyond MMDocIR, we construct an additional evaluation set from UniDoc-Bench~\citep{peng2025unidoc}. This extension is evaluated separately from the 2,024-question main benchmark.

\paragraph{Extension set construction.}
We apply the same construction pipeline, prompts, acceptance thresholds, and validation stages to 60 financial and legal documents from UniDoc-Bench, without corpus-specific modifications. Filename and content-hash checks identify no overlap with MMDocIR. The pipeline produces 5,134 atomic questions and 2,137 bridge-chain candidates. After validation and curation to match the main benchmark's topology distribution, we retain 473 questions: 72 single-hop, 236 bridge-chain, and 165 synthesis questions. The retained instances preserve the same evidence-topology annotations and satisfy the applicable topology-specific validation criteria used for the main benchmark. Table~\ref{tab:unidoc_topology} summarizes the extension, which includes both cross-modal bridge chains and multi-evidence synthesis.

\begin{table}[H]
\centering
\caption{Topology distribution of the 473-question extension constructed from 60 UniDoc-Bench documents. Left: detailed counts. Right: inner sectors show topology proportions, and outer sectors show the corresponding breakdowns and counts. T: text; F: figure; pts: evidence points.}
\label{tab:unidoc_topology}

\begin{minipage}[c]{0.43\linewidth}
\centering
\footnotesize
\setlength{\tabcolsep}{4pt}
\renewcommand{\arraystretch}{1.05}
\begin{tabular}{@{}lr@{\hspace{1em}}lr@{}}
\toprule
\textbf{Topology} & \textbf{Total} &
\textbf{Breakdown} & \textbf{Count} \\
\midrule
\multirow{2}{*}{Single-hop} &
\multirow{2}{*}{72} &
Text & 35 \\
& & Figure & 37 \\
\midrule
\multirow{3}{*}{Bridge-chain} &
\multirow{3}{*}{236} &
Text $\rightarrow$ text & 82 \\
& & Text $\rightarrow$ figure & 68 \\
& & Other paths & 86 \\
\midrule
\multirow{3}{*}{Synthesis} &
\multirow{3}{*}{165} &
3 evidence points & 92 \\
& & 4 evidence points & 64 \\
& & Other & 9 \\
\midrule
\textbf{Total} & \textbf{473} & & \\
\bottomrule
\end{tabular}
\end{minipage}\hfill%
\begin{minipage}[c]{0.55\linewidth}
\centering
\includegraphics[width=\linewidth]{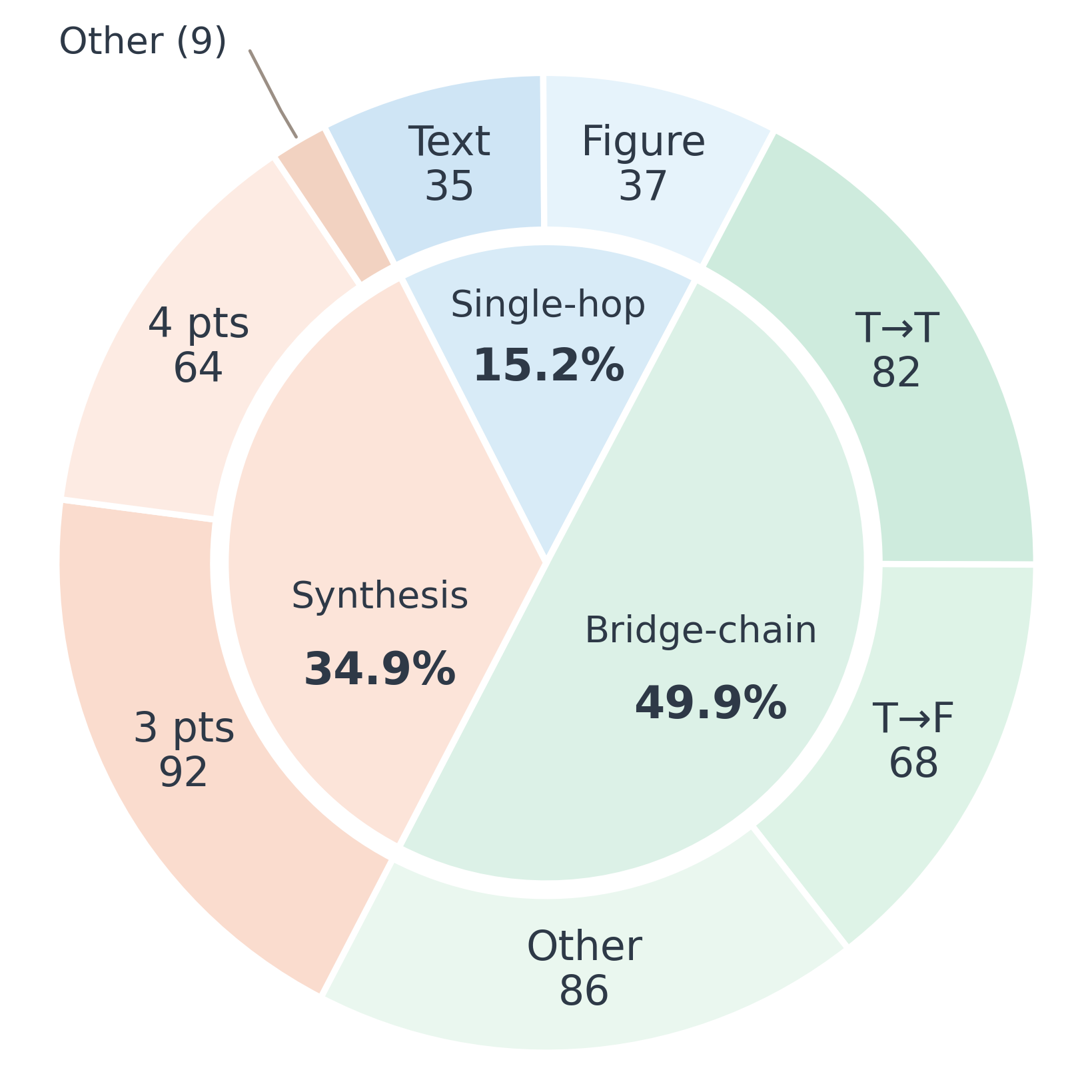}
\end{minipage}

\end{table}

\paragraph{Evaluation setup.}
We evaluate one representative from each retrieval paradigm: LightRAG for text-only GraphRAG, VisRAG for page-level visual retrieval, and RAG-Anything for multimodal GraphRAG. All three systems retain their original configurations and use the same evaluation protocol as the main benchmark, as detailed in Appendices~\ref{app:baseline_config} and~\ref{app:metrics}. The systems are evaluated on all 473 questions. Step coverage is computed only for the 401 bridge-chain and synthesis questions, consistent with its applicability in the main experiments.

\begin{table}[H]
\centering
\caption{Performance on the UniDoc-Bench extension. Step coverage is computed over bridge-chain and synthesis questions only. Text-based context precision, context recall, and faithfulness are not reported for VisRAG because its retrieved contexts are page images. Bold values indicate the highest score among the three evaluated systems.}
\label{tab:unidoc_results}
\footnotesize
\setlength{\tabcolsep}{4pt}
\renewcommand{\arraystretch}{1.08}
\begin{tabular*}{\linewidth}{@{\extracolsep{\fill}}lcccccc@{}}
\toprule
\textbf{System} &
\shortstack{\textbf{Context}\\\textbf{Precision}} &
\shortstack{\textbf{Context}\\\textbf{Recall}} &
\shortstack{\textbf{Answer}\\\textbf{Accuracy}} &
\textbf{Faithfulness} &
\shortstack{\textbf{Response}\\\textbf{Relevancy}} &
\shortstack{\textbf{Step}\\\textbf{Coverage}} \\
\midrule
LightRAG &
0.7324 & 0.6518 & 0.5356 & 0.7871 & 0.6942 & 0.5716 \\
VisRAG &
-- & -- & 0.3800 & -- & 0.7133 & 0.5021 \\
RAG-Anything &
\textbf{0.8473} & \textbf{0.7259} & \textbf{0.5842} &
\textbf{0.8847} & \textbf{0.8159} & \textbf{0.7632} \\
\bottomrule
\end{tabular*}
\end{table}

\paragraph{Results.}
Table~\ref{tab:unidoc_results} shows that RAG-Anything achieves the highest answer accuracy and step coverage among the three systems. Its answer accuracy is 0.5842, compared with 0.5356 for LightRAG and 0.3800 for VisRAG. Its step coverage reaches 0.7632, compared with 0.5716 and 0.5021, respectively. Both metrics therefore reproduce the ordering observed on MMDocIR: RAG-Anything outperforms LightRAG, which in turn outperforms VisRAG.

With the construction procedure, system configurations, and evaluation protocol held fixed across corpora, these results support the applicability of our methodology beyond MMDocIR. They also show that the advantage of multimodal GraphRAG in final-answer correctness and coverage of required reasoning content persists on the extension set for the three representative systems evaluated here.

\section{Limitations}
\label{app:limitations}

This work has several limitations. First, \textsc{TopoGraphRAG-Bench} is built on a sampled subset of MMDocIR, so its document domains, layout styles, and evidence distributions are bounded by the source corpus. Future work could extend the benchmark to broader document collections, including scientific papers, reports, manuals, and real world enterprise documents.

Second, the benchmark focuses on three controlled evidence topologies: single hop retrieval, bridge chain reasoning, and multi source synthesis. These categories enable targeted evaluation of evidence structure recovery, but they do not cover all possible forms of document reasoning or longer dependency chains. Future work could expand the topology space to support more diverse and realistic reasoning patterns.

Third, the evaluated systems represent major retrieval paradigms, including text only GraphRAG, page level visual retrieval, and multimodal GraphRAG, but they do not exhaust all possible parsers, retrievers, graph construction strategies, or generation backends. Our experiments focus on comparing retrieval and indexing paradigms under controlled generation and embedding configurations, rather than systematically assessing sensitivity to backbone choice. In addition, GraphRAG systems can incur substantial offline indexing costs because entities, relations, communities, or multimodal graph structures must be constructed before retrieval. Future work could study additional system designs, quality cost tradeoffs, and incremental indexing strategies for large scale multimodal GraphRAG deployment.

\section{License and Data Use}
\label{app:license}

Our benchmark is built on a sampled subset of the MMDocIR evaluation dataset
\citep{dong2025mmdocir}. We use 201 documents and their associated
layout-grounded document units as the source corpus for constructing new
questions, evidence-topology annotations, and answers. The original documents,
page images, OCR outputs, layout units, and MMDocIR annotations remain subject
to the license and usage terms of the upstream MMDocIR dataset. We do not claim
ownership over these source materials.

The new annotations introduced in this work, including the generated questions,
gold answers, evidence topology labels, bridge entities, synthesis evidence
points, and validation metadata, are released for academic and noncommercial
research use. Users of the benchmark are responsible for complying with the
upstream MMDocIR license and any applicable terms for the original documents.
When redistributing or using this benchmark, users should cite both MMDocIR and
this work.




\end{document}